\documentclass[sigconf]{acmart}

\copyrightyear{2026}
\acmYear{2026}
\setcopyright{cc}
\setcctype{by}
\acmConference[UIST '26]{The 39th Annual ACM Symposium on User Interface Software and Technology}{November 02--05, 2026}{Detroit, MI, USA}
\acmBooktitle{The 39th Annual ACM Symposium on User Interface Software and Technology (UIST '26), November 02--05, 2026, Detroit, MI, USA}
\acmDOI{10.1145/3830398.3830712}
\acmISBN{979-8-4007-2856-3/2026/11}

\usepackage{array}
\usepackage{multirow}
\usepackage{float}
\usepackage{algorithm}
\usepackage{algpseudocode}
\usepackage{colortbl}
\usepackage{tabularx}
\usepackage{booktabs}
\usepackage{enumitem}
\usepackage{framed}
\definecolor{shadecolor}{gray}{0.96}

\newenvironment{promptbox}[2][]{%
  \begin{shaded}\noindent\textbf{\small #2}\par\smallskip\small
}{\end{shaded}}

\newenvironment{wideprompt}{\begin{shaded}}{\end{shaded}}

\usepackage{multicol}   
\usepackage{array}      
\usepackage{pifont}
\usepackage{xurl}
\newcommand{\metric}[1]{\path{#1}}
\newcommand{\cmark}{\ding{51}}
\newcommand{\xmark}{\ding{55}}
\definecolor{thinkcolor}{RGB}{56,189,248}
\definecolor{implcolor}{RGB}{74,222,128}
\definecolor{debugcolor}{RGB}{248,113,113}
\definecolor{seekcolor}{RGB}{167,139,250}
\definecolor{testcolor}{RGB}{251,191,36}

\definecolor{cColdStart}{HTML}{5B8BD4}
\definecolor{cOriented}{HTML}{4AA84F}
\definecolor{cPassive}{HTML}{F0943D}
\definecolor{cIterating}{HTML}{8DC572}
\definecolor{cDebugging}{HTML}{E05A3A}
\definecolor{cSpinning}{HTML}{7B5EA7}
 
\definecolor{cDelegation}{HTML}{7BA3D4}
\definecolor{cTargeted}{HTML}{3D8B40}
\definecolor{cConceptual}{HTML}{E8944A}

\newcommand{\sys}{TutorTrace}

\AtBeginDocument{%
  }

\begin{document}

\title{\sys{}: A Dataset and Taxonomy for Classifying Learner Behavioral States during AI-Assisted Programming Education}

\author{David Barron}
\orcid{0009-0005-7252-7762}
\affiliation{%
  \institution{Virginia Tech}
  \city{Blacksburg}
  \state{Virginia}
  \country{USA}}
\email{dbarron410@vt.edu}

\author{Xiaohang Tang}
\orcid{0000-0002-2691-9280}
\affiliation{%
  \institution{Virginia Tech}
  \city{Blacksburg}
  \state{Virginia}
  \country{USA}}
\email{xiaohangtang@vt.edu}

\author{Rezky Dwisantika}
\orcid{0009-0003-3192-090X}
\affiliation{%
  \institution{Sepuluh Nopember Institute of Technology}
  \city{Surabaya}
  \state{Jawa Timur}
  \country{Indonesia}}
\email{rezkysantika21@gmail.com}

\author{Minsun Kim}
\orcid{0009-0002-4593-2672}
\affiliation{%
  \institution{Virginia Tech}
  \city{Blacksburg}
  \state{Virginia}
  \country{USA}}
\email{minsunkim@vt.edu}

\author{David H. Smith IV}
\orcid{0000-0002-6572-4347}
\affiliation{%
    \institution{Virginia Tech}
  \city{Blacksburg}
  \state{Virginia}
  \country{USA}}
\email{dhsmith4@vt.edu}

\author{Jiaming Cui}
\orcid{0000-0002-2685-2776}
\affiliation{%
    \institution{Virginia Tech}
  \city{Blacksburg}
  \state{Virginia}
  \country{USA}}
\email{jiamingcui@vt.edu}

\author{Yan Chen}
\orcid{0000-0002-1646-6935}
\affiliation{%
    \institution{Virginia Tech}
  \city{Blacksburg}
  \state{Virginia}
  \country{USA}}
\email{ych@vt.edu}

\renewcommand{\shortauthors}{Barron et al.}

\begin{abstract} 
AI programming tutors provide scalable support, yet lack the
behavioral context human tutors rely on to adapt support to
learners' needs. We present \sys{}, a dataset and behavioral
abstraction pipeline that makes learners' behavioral
context visible and computable in real time from low-level IDE
telemetry. Across four deployments in two introductory Python courses (N=480), TutorTrace captures approximately 180K telemetry events, 13,633 behavioral segments, and 27 continuously computed metrics. From this foundation, we derive a taxonomy of learner
activity before the first AI query, between consecutive queries,
and across the full session, enabling systems to respond not just to what learners say, but to what they have done leading up to the help-seeking moment. In a preliminary classroom evaluation, behavior-aware prompts
were associated with a decrease in intervals between queries
with no independent work from 50.0\% to 20.7\%. As an additional demonstration of downstream utility, we
evaluate \sys{} on two held-out prediction tasks: whether a
learner will query within the next 60 seconds (AUROC${=}.726$) and whether an upcoming query reflects guided
or dependent help-seeking (AUROC${=}.717$). Together, these findings show how behavioral context can enable adaptive AI tutoring at scale.
\end{abstract}


\maketitle

\section{Introduction}

\begin{table*}[t]
\centering
\caption{Comparison of \sys{} with existing CS education and AI-assisted
programming datasets: the 2nd CSEDM Data Challenge~\cite{csedm2021challenge, edwards2017codeworkout},
ProgSnap2~\cite{price2020progsnap2}, CodeAid~\cite{kazemitabaar2024codeaid},
and CodeWatcher~\cite{basha2025codewatcher}.}
\label{tab:dataset_comparison}
\renewcommand{\arraystretch}{1.25}
\setlength{\tabcolsep}{5pt}
\small
\begin{tabular}{l ccccc}
\toprule
\textbf{Property}
  & \textbf{CSEDM}
  & \textbf{ProgSnap2$^*$}
  & \textbf{CodeAid}
  & \textbf{CodeWatcher$^\dagger$}
  & \textbf{\sys{} (ours)} \\
\midrule
Real classroom deployment
  & \cmark & \cmark & \cmark & \xmark & \cmark \\
Longitudinal coverage (semester)
  & \cmark & \cmark & \cmark & \xmark & \xmark \\
Raw IDE telemetry
  & \xmark & \cmark & \xmark & \cmark & \cmark \\
AI query \& response logs
  & \xmark & \xmark & \cmark & \xmark & \cmark \\
Task outcome data
  & \cmark & \cmark & \xmark & \xmark & \cmark \\
Behavioral classification (expert-validated)
  & \xmark & \xmark & \xmark & \xmark & \cmark \\
\midrule
$N$ students
  & 819 & varies & 563 & 3$^\dagger$ & 480 \\
$N$ AI interactions
  & --- & --- & 7{,}003 & --- & 1{,}386 \\
\bottomrule
\multicolumn{6}{l}{\footnotesize $^*$ProgSnap2 is a data format specification;
student counts vary by adopting institution.} \\
\multicolumn{6}{l}{\footnotesize $^\dagger$CodeWatcher is a telemetry-collection
tool (demonstration); $N$ reflects an illustrative use case with three
developers, not a released dataset.}
\end{tabular}
\end{table*}

Large Language Models have improved personalized computing education by providing learners with near-instant, in-depth responses built on a knowledge base no single tutor can compete with \cite{tang2024sphere}. Yet, one-on-one human tutoring remains the gold standard for improving learner outcomes \cite{bloom19842}. One key reason is that human tutors respond not only to what learners explicitly say, but also to what their observable behaviors suggest they implicitly need~\cite{wood1976tutoring, chi2001learning}. Consider what this could look like at scale.

\begin{quote}
\addtolength{\leftskip}{-0.2cm}
\addtolength{\rightskip}{-0.2cm}
    {Imagine if every student in a 400-person programming course had a tutor sitting beside them, observing what they do and where they struggle. One who knows when to push them harder and when to slow down and offer support. One who could tell the difference between a student who is struggling and one who hasn't put in the effort. Achieving this vision requires systems that can infer learners' evolving behavioral context in real time from fine-grained interaction data, a capability and data resource not yet available to our community.}
\end{quote}

The human tutor's comparative advantage lies in the behavioral context that precedes the learner's help-seeking moment. Current AI tutoring systems are typically limited to the question itself, missing the struggle or lack thereof that preceded it. To bridge this gap, systems must be able to programmatically detect the behavioral patterns that human tutors observe intuitively.

This requires datasets that expose fine-grained student programming behavior at scale. Most existing datasets capture only subsets of the learner's programming process, such as static code submissions or AI interaction logs. Few provide the continuous, fine-grained interaction telemetry required to enable full-fidelity replays of learner programming sessions (Table~\ref{tab:dataset_comparison}). Systems that do capture interaction at this granularity \cite{mozannar2024reading, zhang2026editrail} were not built to release classroom-scale datasets.

Prior work shows that the behavioral context surrounding a help-seeking moment is associated with learning outcomes \cite{ma2025not}, and specific interaction
patterns distinguish effective from ineffective AI use \cite{ma2025not, shen2026aiimpactsskillformation}. Students who invest independent effort before receiving instruction learn more from it \cite{kapur2014productive, sinha2021productive}, and a student's behavioral context directly informs the tutor's pedagogical strategy for intervening \cite{harvey2025don}. Yet current AI tutoring systems typically respond to the information learners explicitly provide rather than the
behavioral process that preceded it. As a result, two learners
with similar code, errors, and questions may receive similar
support despite arriving at that moment through fundamentally
different processes.

We present \sys{}, a dataset and automated behavioral abstraction infrastructure designed to make this context visible and computable. The dataset comprises fine-grained IDE telemetry from 480 students across four deployments in two introductory Python courses. An automated pipeline transforms this telemetry into continuously computed observable metrics and behavioral
sequences. Unlike existing datasets that capture only what learners submit or ask, \sys{} captures the story of how they got there, making each learner's evolving behavioral context programmatically accessible in real time.

To demonstrate the utility of \sys{} for future research and system design, we examine how its representations of behavioral context can support adaptive AI tutoring. First, we derive a three-window behavioral taxonomy that
organizes learner activity before the first AI query, between
consecutive queries, and through recurring re-querying patterns
across the session. Next, we conduct a preliminary comparison examining whether behavior-aware prompting is associated with changes in learners'
observable activity between queries. Finally, we demonstrate additional predictive utility through
two tasks: predicting whether a learner will initiate an AI
interaction within the next 60 seconds and whether an upcoming
query will reflect guided or dependent help-seeking. Together, these capabilities enable systems to detect, respond to, and anticipate learner behavior surrounding AI interactions. We make the following contributions:

\begin{itemize}[
  topsep=8pt,
  partopsep=0pt,
  parsep=0pt,
  itemsep=4pt
]

\item \textbf{\sys{} Dataset:} A publicly available
dataset\footnote{\url{https://github.com/divadbaroon/TutorTrace_dataset_and_benchmark/tree/uist}} from 480 learners
across four deployments in two introductory Python courses,
comprising approximately 180K fine-grained IDE telemetry events
mapped to 13,633 automatically classified behavioral segments,
along with 27 continuously computed observable metrics
surrounding 1,386 AI interactions.

\item \textbf{\sys{} Classifier:} An automated behavioral
classifier that transforms raw IDE telemetry into labeled
behavioral sequences in real time. The classifier is grounded in expert-developed rules and
validated against expert labels and student self-reports, with
overall pairwise agreement ranging from 78\% to 87\%.

\item \textbf{\sys{} Taxonomy and Downstream Utility:}
We demonstrate how the representations provided by \sys{} can
support adaptive AI tutoring through:
  \begin{itemize}[noitemsep, topsep=2pt]
    \item A three-window behavioral taxonomy comprising ten
    profiles that characterize activity before the first AI
    query, between consecutive queries, and through recurring
    re-querying patterns across the session.

    \item A preliminary classroom comparison in which
    behavior-aware prompting was associated with a decrease in
    Passive inter-query windows from 50.0\% to 20.7\% and
    greater observable activity between queries.

    \item Two held-out prediction tasks demonstrating that
    observable metrics support prediction of query imminence
    within 60 seconds (AUROC${=}.726$) and guided versus
    dependent help-seeking (AUROC${=}.717$).
  \end{itemize}

\end{itemize}

\section{Related Work}

\subsection{Prior Taxonomies in CS Education}
Taxonomies typically organize distinct, non-overlapping categories across one or more dimensions while also revealing relationships among the taxa \cite{irvine2021taxonomies}. Past studies in CS education have employed educational taxonomies to design learning objectives and assess learning effectiveness \cite{smith2025comparing}, though researchers have argued that general-purpose frameworks such as Bloom's taxonomy and SOLO do not fully capture the distinctive characteristics of computer science learning, particularly in programming-related contexts \cite{fuller2007developing}.

After the rise of LLMs, behavioral analysis of programming interaction has become more important and complex. Copilot, an LLM-based code generation tool, can produce correct solutions to many introductory programming problems, raising questions about how its presence reshapes student work \cite{wermelinger2023using}. In CS education specifically, LLM-based programming assistants support students during coding tasks while also reshaping their interaction patterns, problem-solving processes, and reliance on AI assistance \cite{kazemitabaar2024codeaid, liffiton2023codehelp, hou2024codetailor}. As a result, programming sessions generate large amounts of human-AI interaction traces not easily described by traditional educational taxonomies alone \cite{mozannar2024reading, barke2023grounded}. To address this gap, Mozannar et al.~\cite{mozannar2024reading} proposed a taxonomy representing AI-assisted coding sessions as timelines of transitions across programmer activities, while Barke et al.~\cite{barke2023grounded} showed that Copilot use also reflects broader interaction modes, such as acceleration and exploration.

Despite these developments, taxonomies specifically designed to
characterize student behavior in LLM-assisted programming
contexts remain rare. In this work, we introduce a taxonomy of
learner behavioral profiles for AI-assisted programming
education, aiming to make student--AI interaction patterns more
visible and interpretable in educational settings.

\subsection{Help-Seeking Quality and AI Dependency}
With the advancement of AI, students increasingly rely on them to complete assignments rather than as learning aids, raising concerns about over-reliance and reduced independent problem-solving \cite{jovst2024impact, xue2024does, wang2026overreliance}. Recent work in CS education has developed tools that leverage LLMs for scaffolded hint generation while incorporating guardrails to prevent solution delegation \cite{kazemitabaar2024codeaid, hou2024codetailor, liffiton2023codehelp}. However, even when system-level constraints limit the model's output, students bypass them when given the option: Kapoor et al.~\cite{kapoor2026exploring} deployed an AI TA with an optional ``See Solution'' control that disabled the guardrails, and 50\% of 885 students used it on at least one problem, with 14\% using it on all three.

To mitigate these behaviors, several approaches shape or qualify student queries before they reach the underlying LLM. CodeHelp~\cite{liffiton2023codehelp} replaces a single free-form box with structured fields for language, code, error message, and issue description, and runs an LLM-based sufficiency check that returns a clarification request when a query lacks critical information. CodeAid~\cite{kazemitabaar2024codeaid} similarly offers feature-specific input templates that scaffold how students frame a request.
However, these approaches intervene based on the content of the query itself, not the behavioral context that preceded it. A student who spent five minutes debugging independently before asking and a student who re-queried immediately after the last response may submit identical questions, yet require fundamentally different pedagogical responses. \sys{} addresses this gap by making the behavioral context surrounding each query visible and computable, enabling interventions grounded not in what the student says, but in what they did before saying it.

\begin{figure}[t]
  \includegraphics[width=\columnwidth]{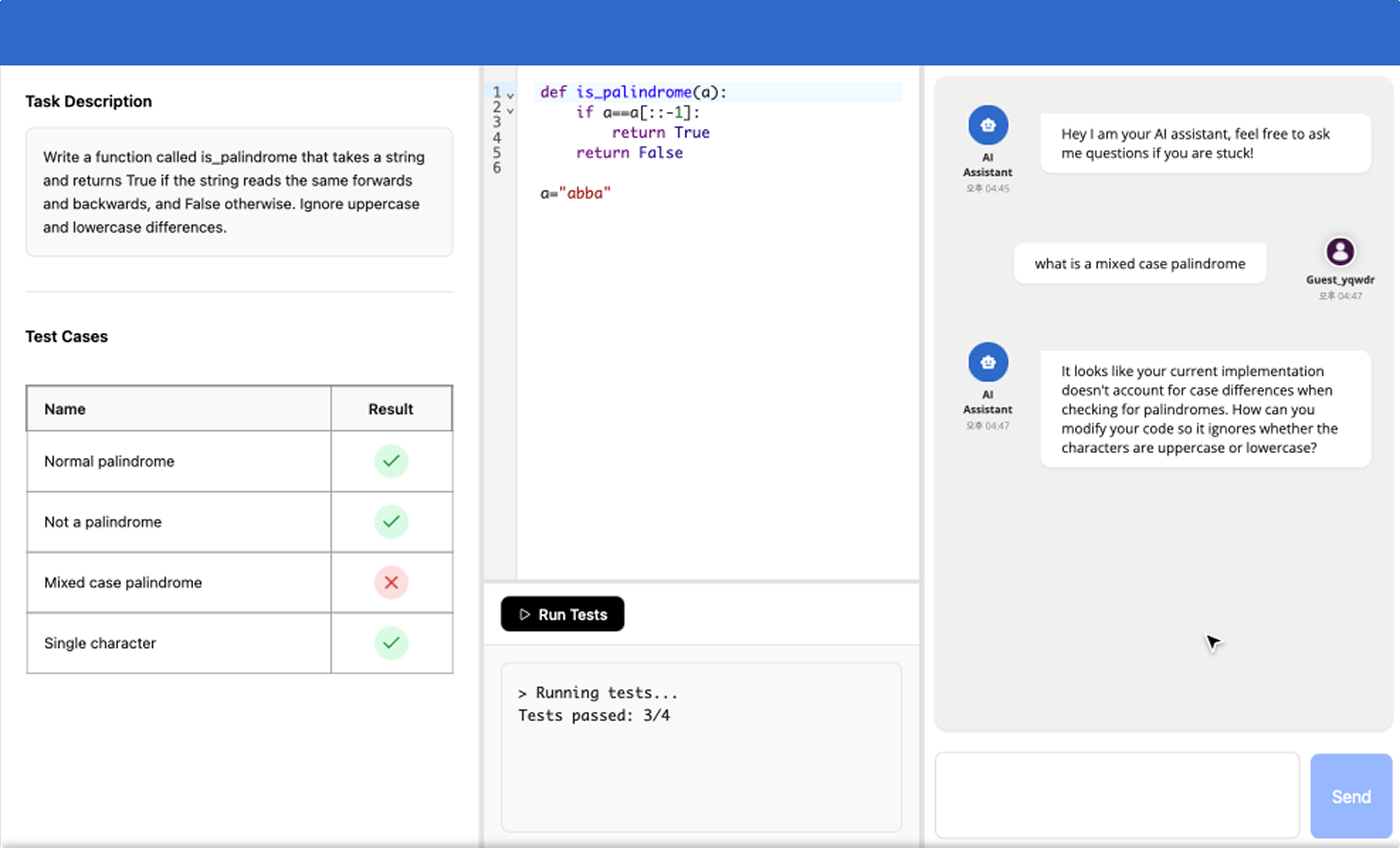}
  \caption{The student interface with task description, 
  code editor, terminal, test cases, and AI chat window.}
  \Description{Screenshot of the student IDE showing a task description and test cases on the left, a code editor and terminal in the center, and an AI chat window on the right.}
  \label{fig:student-ide}
\end{figure}

\section{System}

\sys{} is a task-based IDE platform with LLM support, similar
in structure to LeetCode\footnote{\url{https://leetcode.com/}}
and HackerRank\footnote{\url{https://www.hackerrank.com/}}.These platforms share a fixed layout containing a task description, test cases, code editor, terminal, and AI chat window, constraining student activity to a well-defined interaction space, making fine-grained behavioral observation within this environment both feasible and reproducible. As students complete programming tasks, \sys{} passively captures and translates low-level telemetry into behavioral sequences and observable metrics in real time, providing programmatic access to the behavioral context that human tutors observe intuitively but that existing AI tutoring systems do not see.

\subsection{Telemetry Capture}

\sys{} captures fine-grained user interactions within the interface, recording 37 event types across six source regions (Table~\ref{tab:telemetry}). Events are captured continuously on the client side and batched to the back end every five seconds with no impact to the student's workflow. Each recorded event includes a millisecond timestamp, source region, and payload. A student's full stream of telemetry events enables a full-fidelity replay of their session in the Behavioral Labeling Interface (Appendix~\ref{app:labeling-interface}, Figure~\ref{fig:labeling-interface}) and serves as the
foundational input to our behavioral abstraction pipeline
(Figure~\ref{fig:abstraction}).

\begin{table}[t]
\centering
\caption{Telemetry events captured by \sys{}, grouped by source region.}
\label{tab:telemetry}
\small
\begin{tabular}{llp{3.2cm}}
\toprule
\textbf{Region} & \textbf{Example Events} & \textbf{Example Payload} \\
\midrule
Code Editor & TYPE, DELETE, PASTE, & Characters, cursor \\
(11 types) & COPY, CUT, UNDO, & position, selection \\
           & REDO, SELECT, INDENT & range \\
\midrule
Terminal & RUN, OUTPUT, ERROR, & stdout, error type \\
(6 types) & RESULT, SELECT, COPY & and message, test \\
          &                      & pass/fail counts \\
\midrule
Chat & QUERY, RESPONSE, & Message content, \\
(9 types) & TYPE, DELETE, PASTE, & length, latency, \\
          & SELECT, COPY & sender ID \\
\midrule
Task/Tests & SELECT, COPY & Selected text, \\
(4 types) &              & source region \\
\midrule
Global & MOUSE\_CLICK, & Coordinates, active \\
(5 types) & TAB\_STATE, & region, tab \\
          & WINDOW\_RESIZE, & visibility, panel \\
          & PANEL\_RESIZE, & dimensions \\
          & MOUSE\_MOVE & \\
\midrule
Session & START, END & Window dimensions, \\
(2 types) &           & duration \\
\bottomrule
\end{tabular}
\end{table}

A complete schema of all 37 event types with example payloads is provided in Appendix~\ref{app:schema}, Table~\ref{tab:telemetry-full}

\subsection{LLM Integration}
\label{sec:llm-integration}

Our system is integrated with GPT-4o\footnote{\url{https://openai.com/index/hello-gpt-4o/}}, accessible via the chat window panel. The prompt is grounded in prior literature on pedagogical prompting strategies for AI tutoring and is structured around three escalating levels of scaffolding \cite{wood1976tutoring, kapur2014productive, roll2011improving}. It was iteratively refined through pilot testing until the LLM consistently adhered to the following principles (Appendix~\ref{app:prompt}):

\begin{enumerate}
    \item \textbf{Socratic Questioning:} Guide the student toward the answer through targeted questions rather than direct explanation. Avoid revealing solutions~\cite{kazemitabaar2024codeaid}.
    \item \textbf{Conceptual Hint:} Provide a high-level conceptual nudge identifying the relevant concept or approach without specifying implementation~\cite{liffiton2023codehelp}.
    \item \textbf{Concrete Scaffolding:} Provide direct, specific guidance using pseudocode and blanks when the student has demonstrated sustained effort without progress~\cite{kapur2014productive}.
\end{enumerate}

\section{Behavioral Classification}

While prior work often derives behavioral labels from heuristic rules alone, we ground ours in the judgment of four domain-expert annotators, each with teaching and research experience in CS education. To label learner sessions, annotators used the Behavioral Labeling Interface (Appendix~\ref{app:labeling-interface}), which presents a replay of the learner’s session, a session timeline, and a behavioral codebook for annotating observed behavior directly onto the timeline. Annotators independently coded replays of learner sessions and met weekly to reconcile disagreements in their classifications.

\subsection{Codebook Development}
\label{sec:codebook-development}

We derived our initial behavioral categories from prior literature on programming behavior and help-seeking~\cite{mozannar2024reading, blikstein2011learning}. We iteratively refined the codebook through 10 pilot labeling sessions drawn from a preliminary deployment in an intermediate Python course at our institution. During this phase, four domain-expert annotators independently segmented and labeled 10 pilot sessions (each 5--13 minutes of learner activity, mean 9 minutes) using the Behavioral Labeling Interface (Appendix~\ref{app:labeling-interface}). The annotators met weekly to reconcile disagreements over segment boundaries and behavioral classifications, updating the codebook as needed (Appendix~\ref{app:codebook-history},
Table~\ref{tab:codebook-history}). The finalized codebook, presented in Table~\ref{tab:codebook}, served as the basis for the segmentation and classification rules used by our automated behavioral classification pipeline.

\begin{table}[h]
\centering
\caption{Behavioral codebook used by domain-expert annotators to label learner sessions. The table lists each behavior, its subtypes and definition, and the corresponding automated segmentation and classification rule derived through annotator consensus except where noted.}
\label{tab:codebook}
\small
\begin{tabular}{p{1.3cm}p{1.4cm}p{1.6cm}p{2.9cm}}
\toprule
\textbf{Behavior} & \textbf{Subtype} & \textbf{Definition} & \textbf{Derived rule} \\
\midrule
Implementing
  & ---
  & Writing new code
  & \texttt{CODE\_EVENT} with no unresolved terminal error \\
\midrule
Debugging
  & ---
  & Fixing an error
  & \texttt{CODE\_EVENT}s while an unresolved error or failed test exists \\
\midrule
Testing\textsuperscript{$\dagger$}
  & ---
  & Executing tests
  & Automatically marked at each \texttt{TERMINAL\_RUN} event \\
\midrule
\multirow{5}{1.3cm}{Thinking}
  & Task      & Reading the task   & $\geq$3s inactivity before any CODE, TERMINAL, or CHAT EVENT \\
\cmidrule(l){2-4}
  & Code      & Reviewing code     & $\geq$6s gap between code edit events \\
\cmidrule(l){2-4}
  & Error      & Reading an error   & $\geq$3s inactivity after \texttt{TERMINAL\_ERROR} \\
\cmidrule(l){2-4}
  & Pre-query  & Formulating query  & $\leq$5s pause before \texttt{CHAT\_QUERY} (stored as metadata) \\
\cmidrule(l){2-4}
  & Response  & Reading AI reply   & \texttt{CHAT\_RESPONSE} when the preceding segment was chat input \\
\midrule
Seeking Help
  & ---
  & Typing query to AI
  & \texttt{CHAT\_TYPE}, \texttt{CHAT\_QUERY} events \\
\midrule
Idle
  & ---
  & No activity
  & No events of any type for $>$15s \\
\midrule
Off-Topic 
  & ---
  & Unrelated activity
  & Input unrelated to task \\
\midrule
Unknown 
  & ---
  & Cannot classify
  & Unknown telemetry event \\
\bottomrule
\end{tabular}

\vspace{3pt}
\begin{minipage}{\linewidth}
\footnotesize
\textsuperscript{$\dagger$} \textit{Testing was not manually annotated. Because test execution is directly observable, each \texttt{TERMINAL\_RUN} event was automatically inserted as a run marker on the Behavioral Labeling Interface timeline.}
\end{minipage}
\end{table}

\subsection{Behavioral Segmentation}

The most challenging aspect of manual annotation was determining precise segment boundaries. Across the pilot sessions, annotators generally agreed on which behavior was occurring but differed in the exact timing of its start and end. The mean discrepancy in boundary placement across annotators was 2.3 seconds. During weekly reconciliation meetings, we found this stemmed from the cognitive load of continuously watching session replays rather than any conceptual disagreement on what constituted a behavioral transition. Annotators reported that identifying the behavior itself was straightforward, whereas pinpointing the exact time of transition required sustained attention that naturally degraded over longer sessions.

To address this, the first author reviewed all independently annotated boundaries and synthesized them into proposed consensus segment boundaries. To verify these, all annotators then rewatched each session while the first author verbally announced each boundary as it occurred. This reconciliation process yielded consensus on all segment boundaries. We used the resulting consensus boundaries to derive the automated segmentation rules described in Appendix~\ref{app:algorithm},
Algorithm~\ref{alg:autoseg}.

\begin{table}[t]
\vspace{-8pt}
\centering
\caption{Pairwise agreement among learner self-reports, expert annotations, and the automated classifier across six user-study sessions. Cohen's $\kappa$~\cite{cohen1960kappa} and raw percentage agreement are reported for each behavior and overall.}
\label{tab:irr-threeway}
\small
\setlength{\tabcolsep}{3pt}
\renewcommand{\arraystretch}{1.2}
\begin{tabular}{l rr rr rr}
\toprule
& \multicolumn{2}{c}{\textbf{Self vs Expert}}
& \multicolumn{2}{c}{\textbf{Expert vs Auto}}
& \multicolumn{2}{c}{\textbf{Self vs Auto}} \\
\cmidrule(lr){2-3}\cmidrule(lr){4-5}\cmidrule(lr){6-7}
\textbf{Behavior}
& $\kappa$ & Agr.
& $\kappa$ & Agr.
& $\kappa$ & Agr. \\
\midrule
Implementing & 0.93 & 97\% & 0.95 & 98\% & 0.98 & 99\% \\
Debugging & 0.76 & 96\% & 0.88 & 98\% & 0.71 & 94\% \\
Seeking Help & 0.90 & 98\% & 1.00 & 100\% & 0.90 & 98\% \\
\midrule
\textit{Thinking:} & & & & & & \\
\quad Task & 0.92 & 96\% & 0.96 & 98\% & 0.95 & 98\% \\
\quad Code & 0.64 & 87\% & 0.34 & 82\% & 0.32 & 78\% \\
\quad Error & 0.54 & 87\% & 0.42 & 84\% & 0.42 & 79\% \\
\quad Pre-query & 0.50 & 97\% & 1.00 & 100\% & 0.50 & 97\% \\
\quad Response & 0.50 & 89\% & 1.00 & 100\% & 0.50 & 89\% \\
\midrule
Overall & 0.79 & 83\% & 0.83 & 87\% & 0.73 & 78\% \\
\bottomrule
\end{tabular}
\end{table}

\subsection{Validation}

We validated the pipeline in two stages. First, we established inter-rater reliability among human annotators across the 10 pilot sessions to confirm the codebook was sufficiently stable before scaling. Second, we conducted a user study with 13 participants who completed two Python programming tasks and then self-labeled their behavior using the Behavioral Labeling Interface. Our domain-expert annotators independently labeled six of these sessions, allowing us to triangulate agreement across three sources: learner self-reports, expert annotations, and the automated classifier. Testing was excluded from the agreement analysis because test executions were directly observed from \texttt{TERMINAL\_RUN} events rather than independently annotated. Across the three pairwise comparisons, overall Cohen's $\kappa$ ranged from 0.73 to 0.83, and overall raw agreement ranged from 78\% to 87\% (Table~\ref{tab:irr-threeway}).

We also found several recurring edge cases that raw IDE telemetry could not differentiate alone. For instance, after receiving an error, a learner might ignore it and continue implementing a separate feature, making \textit{Implementing} difficult to distinguish from \textit{Debugging}. In the same vein, \textit{Thinking about Code} may be difficult to distinguish from \textit{Thinking about Error}. For these situations, semantic understanding is needed to differentiate the behavior. We discuss further in Section~\ref{sec:limitations}.

\begin{table}[!t]
\centering
\caption{The 35 candidate observable metrics grouped by observable area, drawn from prior IDE-based learning analytics literature~\cite{blikstein2011learning,mozannar2024reading}. Metrics marked $\dagger$ were excluded during pruning.}
\label{tab:features}
\footnotesize
\setlength{\tabcolsep}{3pt}
\begin{tabularx}{\columnwidth}{>{\raggedright\arraybackslash}p{1.8cm} X X}
\toprule
\textbf{Observable Area} & \multicolumn{2}{l}{\textbf{Metrics}} \\
\midrule
Code activity
& \texttt{code\_edits}         & \texttt{code\_edit\_rate} \\
& \texttt{chars\_inserted}     & \texttt{chars\_deleted} \\
& \texttt{code\_deletes}       & \texttt{net\_code\_growth} \\
& \texttt{delete\_type\_ratio} & \texttt{code\_pastes}\textsuperscript{\dag} \\
\midrule
Terminal activity
& \texttt{terminal\_runs}           & \texttt{terminal\_errors} \\
& \texttt{max\_consecutive\_errors} & \texttt{mean\_time\_between\_runs\_s} \\
\midrule
Error recovery
& \texttt{error\_self\_fix}        & \texttt{error\_reading\_time\_s} \\
& \texttt{error\_to\_edit\_s}      & \texttt{failed\_test\_to\_edit\_s} \\
& \texttt{failed\_test\_self\_fix} & \\
\midrule
Time distribution
& \texttt{time\_in\_editor\_s}  & \texttt{time\_in\_terminal\_s} \\
& \texttt{time\_in\_chat\_s}    & \texttt{time\_in\_task\_s} \\
& \texttt{time\_in\_tests\_s}   & \texttt{longest\_idle\_s} \\
& \texttt{tab\_hidden\_time\_s}\textsuperscript{\dag} & \\
\midrule
Chat behavior
& \texttt{thinking\_time\_s}           & \texttt{seeking\_help\_time\_s} \\
& \texttt{duration\_s}                 & \texttt{response\_reading\_time\_s} \\
& \texttt{chat\_to\_code\_latency\_s}  & \\
\midrule
Interface events
& \texttt{tab\_switches}\textsuperscript{\dag}
& \texttt{copy\_events}\textsuperscript{\dag} \\
& \texttt{paste\_events}\textsuperscript{\dag}
& \texttt{undo\_events}\textsuperscript{\dag} \\
& \texttt{redo\_events}\textsuperscript{\dag}
& \texttt{select\_events}\textsuperscript{\dag} \\
\bottomrule
\end{tabularx}

\vspace{2pt}
\vspace{2pt}
{\footnotesize $^\dagger$ Excluded during pruning: sparse metrics ($>80\%$ zeros) and metrics capturing activity outside the instrumented workspace.}
\end{table}

\section{Observable Metrics}
\label{sec:quantification}

While behavioral sequences capture the temporal progression of learner behavior,
they do not provide insights into the learners' aggregate activity within these temporal windows, such as the frequency of code edits, terminal runs, and character deletions. To capture this aggregate activity, we compute observable metrics that convert raw IDE telemetry into intuitive measures of the amount, frequency, and distribution of learner activity within a specific temporal window. Together, behavioral sequences and observable metrics provide complementary perspectives on the temporal evolution and aggregate characteristics of learner behavior within the programming environment.

We compute 35 candidate observable metrics drawn from prior IDE-based learning analytics literature (Table~\ref{tab:features}). These metrics capture complementary dimensions of code activity, terminal activity, error recovery, time distribution, chat behavior, and interface events. After excluding sparse metrics ($>80\%$ zeros) and metrics capturing activity outside the instrumented workspace, 27 observable metrics remain. Together, the retained metrics provide general-purpose aggregate representations of learner activity.

\begin{figure*}[t]
\centering
\vspace{8pt}
\includegraphics[width=1.0\textwidth]{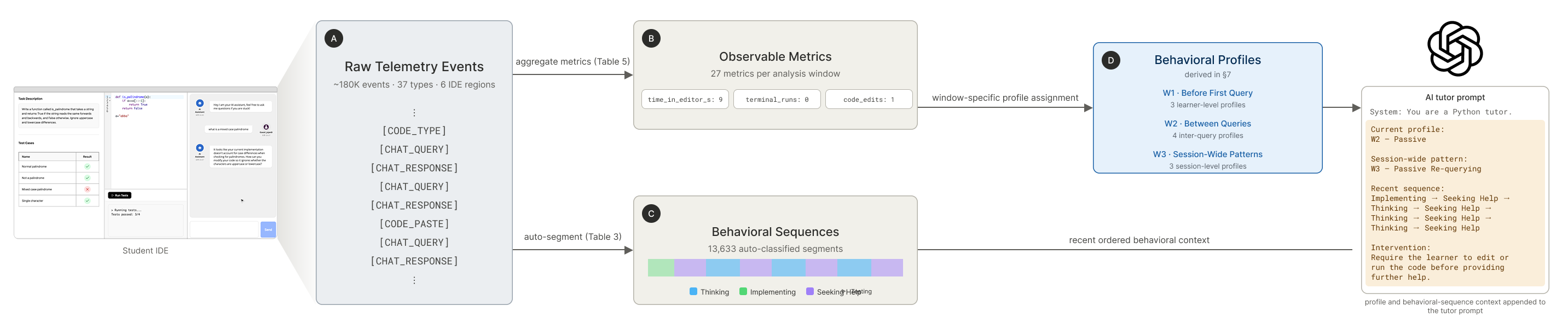}
\caption{\sys{} behavioral abstraction pipeline.
Raw telemetry events (A) are aggregated into window-specific
observable metrics (B) and auto-segmented into behavioral
sequences (C). Observable metrics support assignment to the
behavioral profiles derived in the
\Description{Pipeline diagram. Telemetry from the student IDE streams into raw event tokens, which are aggregated into per-window observable metrics and auto-segmented into behavioral sequences. Metrics feed profile assignment in a unified Behavioral Profiles card listing three, four, and three profiles for Windows 1, 2, and 3, while the sequence feeds the prompt directly. An AI tutor prompt box shows the profile, session pattern, recent sequence, and intervention appended after the system line.}
\sys{} Taxonomy (D), while
behavioral sequences preserve recent ordered activity. These
representations can be incorporated into an AI tutor prompt to
support behavior-aware adaptation.}
\label{fig:abstraction}
\end{figure*}

\section{Deployments and Dataset}

We deployed \sys{} across four deployments in two introductory Python courses at our institution (Table~\ref{tab:deployment}). Deployments 1 and 2 were conducted with Instructor A using a task focused on list indexing, slicing, and in-place manipulation with a 15-minute time limit. Deployments 3 and 4 were conducted with Instructor B using a task focused on nested lists and iteration. Full task descriptions and test cases are provided in Appendix~\ref{app:tasks}. Due to scheduling constraints in the second course with Instructor B, students in Deployments 3 and 4 completed the task under a 10-minute time limit rather than the 15 minutes used in Deployments 1 and 2, which may have contributed to the lower completion rates in those sessions.

The predictive models reported in Section~\ref{sec:predpower} were trained on Deployment 1 (Instructor A, morning, $n=190$) and evaluated on held-out Deployment 2 (Instructor A, afternoon, $n=113$), providing a cross-cohort evaluation under the same instructor and task but at a different time of day. Deployments 3 and 4 supported a preliminary classroom evaluation examining whether behaviorally aware prompts could support changes in learner behavior at scale (Section~\ref{sec:prelim}). Deployment 3 (Instructor B, morning, $n=70$) served as the baseline condition, while Deployment 4 (Instructor B, afternoon, $n=107$) served as the intervention condition, in which the AI tutor's prompts were informed by each learner's behavioral profile. 

\begin{table}[h]
\centering
\caption{Deployment context across four sessions.}
\label{tab:deployment}
\small
\begin{tabular}{lrrrr}
\toprule
\textbf{Deployment}    & \textbf{1} & \textbf{2} & \textbf{3} & \textbf{4} \\
\midrule
Instructor             & A        & A        & B        & B \\
Time limit             & 15 min   & 15 min   & 10 min   & 10 min \\
Task                   & Playlist & Playlist 
& Grade Book & Grade Book \\
Total students         & 190      & 113      & 70       & 107 \\
Used AI                & 94       & 90       & 48       & 85 \\
AI Interactions         & 428      & 540      & 190      & 228 \\
Task completion        & 87.2\%   & 82.3\%   & 34.7\%   & 44.3\% \\
\bottomrule
\end{tabular}
\end{table}

Together, these deployments yield the \sys{} foundation dataset, comprising approximately 180K raw telemetry events, 27 continuously computed observable metrics, and 13,633 automatically classified behavioral sequences (Layers A--C in Figure~\ref{fig:abstraction}).

\section{The \sys{} Taxonomy}
\label{sec:taxonomy}

While our real-time behavioral abstraction pipeline enables systems to be behaviorally aware, making this awareness actionable requires identifying the recurring behavioral signatures that are both distinct and associated with differences in task outcomes. We're interested in both distinct behavioral signatures and task outcomes because learners who are performing poorly with high effort, and those that are performing well with low effort, each require fundamentally different pedagogical responses. We therefore derive the \sys{} Taxonomy, which 
organizes learner activity surrounding AI help-seeking into recurring behavioral profiles, their respective outcomes, and expert-informed interventions that are provided to the AI tutor's prompt.

\begin{figure*}[t]
\centering
\vspace{8pt}
\includegraphics[width=0.96\textwidth]{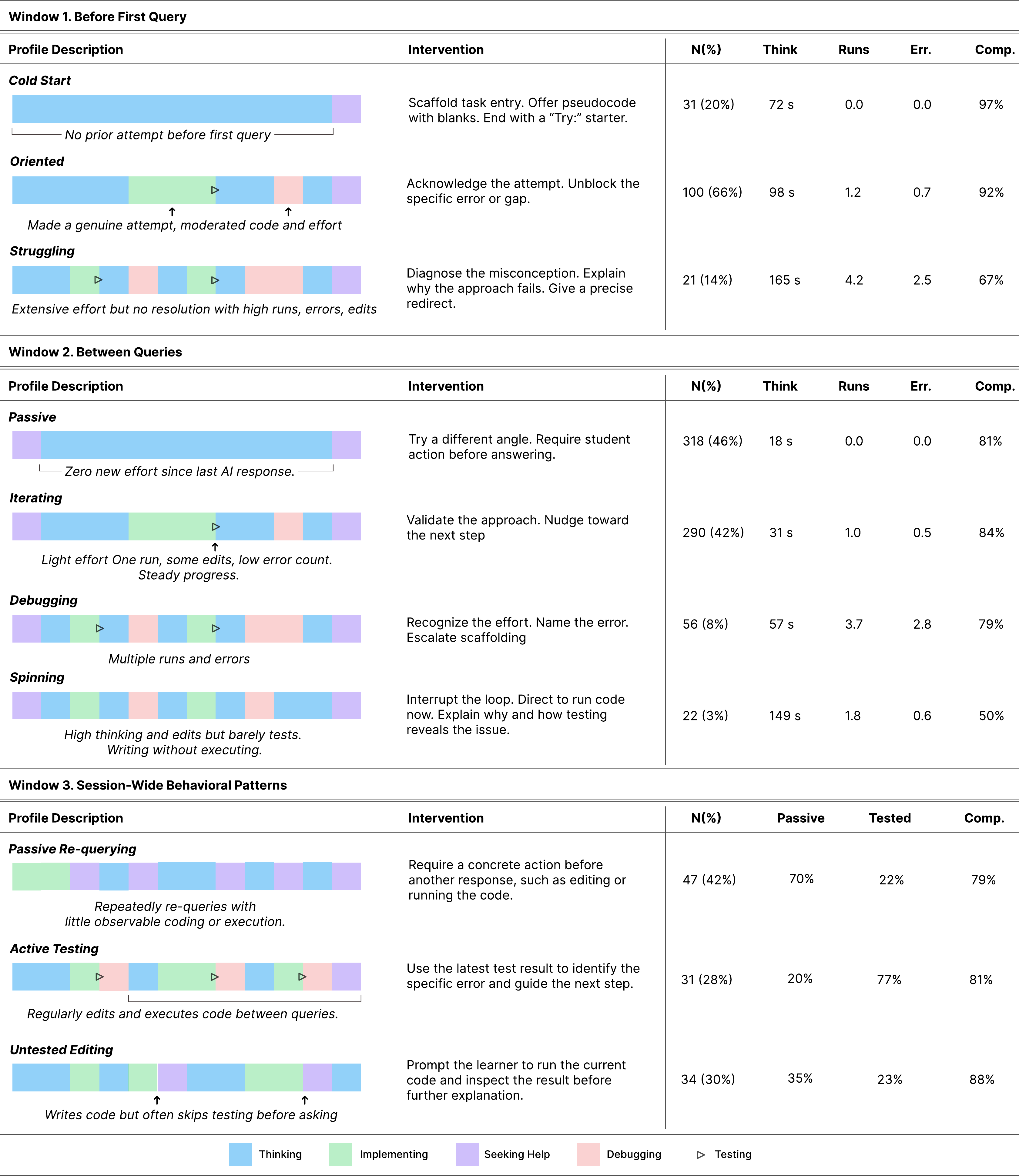}
\caption{The \sys{} three-window behavioral taxonomy. Each window captures a distinct moment in the help-seeking cycle, characterized by different behavioral signals and requiring different intervention strategies.}
\Description{Table of the three-window taxonomy: each row shows a profile with an example behavioral sequence strip, its description, intervention, share of observations, descriptive statistics, and completion rate.}
\label{fig:taxonomy-unified}
\end{figure*}

\subsection{Taxonomy Construction}

We organize the analysis across three temporal windows, each
capturing a distinct stage of the help-seeking cycle: learner
activity before the first AI query (Window 1), activity between
consecutive queries (Window 2), and recurring inter-query
patterns across the session (Window 3). We truncate each session at its first all-pass test result,
which marks the end of task-directed activity and prevents
post-completion behavior from influencing the profiles. For
learners who do not complete the task, we retain activity
through the observed session end. We also exclude individual windows containing more than
30 seconds of tab-hidden time to limit the influence
of unobserved off-platform activity 

Windows~1 and 2 follow an outcome-guided clustering procedure
intended to identify behaviorally distinct profiles that also
differ meaningfully in task outcomes. We first separate
observations with no observable programming effort, defined as
zero code edits and zero terminal runs, into rule-defined
profiles. These observations represent a qualitatively distinct
state rather than simply a low-activity version of active
behavior, and including them in K-means could cause the
zero-inflated observations to dominate the resulting clusters.

For the remaining active observations, we construct a
window-specific candidate metric pool so that clustering uses
only measures meaningful within that stage of the help-seeking
cycle. We exclude metrics that are inapplicable to the window or
zero-valued in more than 80\% of observations because they
provide little discriminative information and may produce
unstable or artifact-driven clusters. 

We exhaustively evaluate every three-metric combination from
the resulting candidate pool. For each combination, we
standardize the metrics and apply K-means~\cite{macqueen1967some} with
$K \in \{2,3\}$ using 10 initializations. We discard solutions
containing fewer than 15 observations in any cluster. For each
remaining solution, we calculate its silhouette
score~\cite{rousseeuw1987silhouettes} and the spread between
its highest and lowest task-completion rates. Candidate
solutions are ranked according to:
\[
    \text{score}
    =
    \text{silhouette}
    \times
    \text{completion-rate spread}.
\]
We perform clustering using behavioral metrics only. Completion
rates are used to select among candidate solutions that produce
behaviorally distinct groups. Because completion informs the
selection of the final solution, the resulting differences in
completion rates across clusters are descriptive rather than
confirmatory. After selecting a solution, we assign profile
names based on the behavioral patterns represented by each
cluster centroid.

\subsubsection{Window 1: Before the First Query}

In Window 1 each learner contributes one set of aggregate metrics capturing their activity before their first query.
Learners with zero code edits and zero terminal runs during this
period are assigned to the rule-defined Cold Start profile
($n{=}31$). The remaining 121 active learners are included in
the exhaustive search.

The highest-ranked solution uses
\metric{time_in_editor_s},
\metric{time_in_terminal_s}, and
\metric{time_in_chat_s} at $K{=}2$
($s{=}0.420$; completion-rate spread${=}25.3$ percentage
points). Based on their behavioral centroids, the resulting
clusters are labeled Oriented ($n{=}100$) and Struggling
($n{=}21$). Together with Cold Start, these profiles
distinguish initial help-seeking that follows no observable
programming attempt, a moderate active attempt, or an extended
period of unsuccessful effort.

\subsubsection{Window 2: Between Consecutive Queries}

In Window 2, each interval between consecutive AI queries contributes one set of aggregate metrics capturing the learner’s activity after one AI response and before their next query. Learners may therefore contribute multiple Window 2 observations and may exhibit different profiles at different points in a session.

Inter-query windows with zero code edits and zero terminal runs
following the preceding AI response are assigned to the
rule-defined Passive profile ($n{=}318$). We combine eligible
post-response and subsequent-effort metrics into one candidate
pool and apply the shared exhaustive search to the remaining
368 active windows. Each interval inherits the task-completion
outcome of its corresponding session.

The highest-ranked solution uses
\metric{time_in_editor_s},
\metric{thinking_time_s}, and
\metric{error_self_fix} at $K{=}3$
($s{=}0.547$; completion-rate spread${=}33.8$ percentage
points). Based on their centroids, the resulting clusters are
labeled Iterating ($n{=}290$), Debugging ($n{=}56$), and
Spinning ($n{=}22$). Together with Passive, these profiles
distinguish repeated queries made without observable
programming activity from queries following brief iteration,
active error recovery, or prolonged activity with limited
execution.

\subsubsection{Window 3: Session-Wide Re-querying Patterns}

Whereas Window 2 characterizes individual inter-query moments,
Window 3 summarizes whether particular forms of inter-query
behavior recur across the session. We aggregate each learner's
valid Window 2 intervals and retain learners with at least two
such intervals, ensuring that the representation captures a
recurring pattern rather than a single observation. This yields
112 eligible learners.

We assign each inter-query interval to one of three mutually
exclusive behavioral categories: Passive, containing no code
edits or terminal runs; Tested, containing at least one
terminal run; or Active Untested, containing code edits but no
terminal run. We then represent each learner using four
session-level measures: the proportion of Passive intervals,
the proportion of Tested intervals, the proportion of Active
Untested intervals, and the longest consecutive Passive streak
normalized by the learner's number of valid intervals.

We standardize these four measures and apply K-means for
$K \in \{2,3,4,5,6\}$ using 100 initializations. As in the
preceding windows, we exclude solutions containing fewer than
15 learners in any cluster. Unlike Windows 1 and 2, task
completion is not used for either clustering or model
selection; we select the eligible solution with the highest
silhouette score.

The highest silhouette score occurs at $K{=}3$
($s{=}0.399$), producing the Passive Re-querying
($n{=}47$), Active Testing ($n{=}31$), and Untested Editing
($n{=}34$) profiles. The assignments are identical across
20 random seeds (ARI~\cite{hubert1985comparing}${=}1.000$).

\subsection{Characteristics of Behavioral Profiles}

The three windows capture complementary aspects of AI
help-seeking: activity before a learner's first query, activity
after an AI response and before the learner's next query, and
recurring re-querying patterns across the session. Because task
completion informed the exploratory selection of the Window 1 and Window 2 clustering solutions, their completion rates are reported as descriptive characteristics rather than as
independent evidence of profile validity. Task completion did
not inform the construction or selection of the Window 3
profiles; its completion rates are likewise reported
descriptively.

\paragraph{Window 1: Before the First Query.}

Cold Start learners comprised 20\% of the Window 1 population
($n{=}31$). These learners made no code edits or terminal runs
before their first query, although their pre-query periods
averaged 72 seconds. The corresponding sessions had a 97\%
completion rate. Oriented learners ($n{=}100$, 66\%) exhibited moderate pre-query activity and had a 92\% completion rate. In
contrast, Struggling learners ($n{=}21$, 14\%) spent
substantially longer before querying, averaging 165 seconds,
4.2 terminal runs, and 2.5 errors. Their corresponding sessions had a 67\% completion rate.

These profiles complicate the assumption that
greater pre-query effort necessarily signals better progress.
In which, Cold Start learners completed the task at high rates despite
making no observable programming attempt, whereas Struggling
learners invested the most time and encountered the most errors
but completed at the lowest rate. Rather than implying that low
effort is beneficial, this contrast suggests different
pedagogical needs: Cold Start learners may require prompting to
engage independently, while Struggling learners may have
exhausted their current strategies and require more direct
scaffolding. An adaptive tutor should therefore respond not only
to the amount of prior activity, but also to what that activity
indicates about the learner's progress.

\paragraph{Window 2: Between Queries.}

Passive behavior accounted for 46\% of valid inter-query
windows ($n{=}318$). In these windows, learners issued another
query without editing or executing their code after receiving the preceding AI response. The corresponding sessions had an 81\% completion rate.

Among active windows, Iterating was the most common profile
($n{=}290$, 42\%). These windows involved relatively brief
thinking periods and modest execution activity, and their
corresponding sessions had an 84\% completion rate. Debugging
windows ($n{=}56$, 8\%) involved more frequent execution and
error recovery, averaging 3.7 terminal runs and 2.8 errors.
Their corresponding sessions had a 79\% completion rate.
Spinning windows ($n{=}22$, 3\%) involved substantially longer
thinking periods, averaging 149 seconds, but comparatively
limited execution activity. Their corresponding sessions had
the lowest completion rate among the Window 2 profiles at 50\%.

Because learners may contribute multiple inter-query windows,
these percentages characterize help-seeking moments rather than
fixed learner types. Together, the profiles distinguish
repeated queries made without observable programming activity
from queries following iteration, active debugging, or
prolonged activity with limited execution. These contexts could
help an adaptive tutor determine whether to prompt independent
work, support error recovery, or help a learner move beyond an
unproductive pattern.

\paragraph{Window 3: Session-Wide Re-querying Patterns.}

Passive Re-querying learners comprised 42\% of the Window 3
population ($n{=}47$). On average, 70\% of their valid
inter-query intervals contained no code edits or terminal runs,
22\% included code execution, and 8\% included editing without
execution. The corresponding sessions had a 79\% completion
rate.

Active Testing learners comprised 28\% of the population
($n{=}31$). Their sessions were characterized by regular code
execution: on average, 77\% of their inter-query intervals
included at least one terminal run, while 20\% were Passive and
4\% involved editing without execution. The corresponding
sessions had an 81\% completion rate.

Untested Editing learners comprised the remaining 30\%
($n{=}34$). On average, 43\% of their inter-query intervals
included code edits without a subsequent terminal run, compared
with 35\% Passive intervals and 23\% Tested intervals. The
corresponding sessions had an 88\% completion rate.

Whereas Window 2 characterizes behavior within individual
inter-query intervals, Window 3 summarizes which forms of
inter-query activity recur across a learner's session. Together, the three windows provide complementary context about when learners request AI assistance, what they do between
requests, and whether those behaviors persist across the
session.

\subsection{Preliminary Evaluation}
\label{sec:prelim}

\sys{} enables systems to detect and understand behavioral context in real
time. The natural next question is whether acting on this
information is associated with changes in learners' observable
behavior. We therefore deployed the baseline system in a
morning session (Deployment~3, $n{=}48$ eligible AI users) and
a treatment version in an afternoon session
(Deployment~4, $n{=}85$ eligible AI users) for the same course
and task.

In the treatment condition, the system appended a
behavior-aware intervention block selected from the learner's
recent activity to the AI tutor's prompt at each interaction
(Appendix~\ref{app:profiles}). We subsequently applied
the revised Window~2 taxonomy to both deployments to examine
whether their profile distributions and observable activity
differed. This preliminary evaluation does not measure learning
gains or establish a causal effect.

Passive inter-query windows accounted for 50.0\% of valid
windows in the baseline deployment and 20.7\% in the
intervention deployment ($-$29.3\,pp). Iterating windows
accounted for 39.1\% and 56.0\%, respectively
($+$16.9\,pp). The overall Window~2 profile distribution
differed between deployments
($\chi^2(3){=}27.55$, $p{<}.0001$). Because learners could
contribute multiple windows, we also compared each learner's
proportion of Passive windows. This learner-level comparison
was likewise significant ($U{=}1533$, $p{<}.0001$).

The intervention deployment also exhibited more code edits per
valid inter-query window (23.7 vs.\ 11.7), more terminal runs
(2.1 vs.\ 1.0), and longer periods of activity between queries
(74.9\,s vs.\ 47.0\,s). Completion among eligible AI users was
33.3\% in the baseline deployment and 43.5\% in the
intervention deployment. This difference was not statistically
significant ($p{=}.273$) and remains confounded by the
between-session design.

These exploratory findings indicate that behavior-aware
prompting was associated with less Passive re-querying and
greater observable activity between queries. However, the
nonrandomized, between-session design does not establish that
the intervention caused these differences.

\begin{table*}[t]
\caption{Held-out AUROC for query imminence (query within 60 seconds) and help-seeking type (guided vs.\ dependent within 15 seconds) across nested feature representations. Models were trained on Deployment~1 and evaluated on Deployment~2 ($n{=}113$). $\Delta$ is relative to raw telemetry; bold indicates the best result.}
\label{tab:ablation}
\centering
\setlength{\tabcolsep}{6pt}
\small
\begin{tabular}{@{}l p{4.2cm} *{2}{c} *{2}{c} *{2}{c}@{}}
\toprule
 & & \multicolumn{2}{c}{\textbf{Raw Telemetry}}
 & \multicolumn{2}{c}{\textbf{+Observable Metrics}}
 & \multicolumn{2}{c}{\textbf{+Behavioral Sequences}} \\
\cmidrule(lr){3-4}
\cmidrule(lr){5-6}
\cmidrule(lr){7-8}
\textbf{Prediction Task}
 & \textbf{Description}
 & AUROC & $\Delta$
 & AUROC & $\Delta$
 & AUROC & $\Delta$ \\
\midrule

\textbf{Query imminence}
 & Will the learner submit an AI query within the next
 60 seconds?
 & 0.689 & ---
 & \textbf{0.726} & +0.037
 & 0.719 & +0.030 \\

\midrule

\textbf{Help-seeking type}
 & Will the learner's upcoming query reflect guided or
 dependent help-seeking?
 & 0.690 & ---
 & \textbf{0.717} & +0.027
 & 0.705 & +0.015 \\

\bottomrule
\end{tabular}
\end{table*}

\begin{table}[t]
\centering
\caption{Guided and dependent help-seeking by behavioral profile. Percentages are calculated within profiles.}
\label{tab:profile-query}
\small
\setlength{\tabcolsep}{5pt}
\renewcommand{\arraystretch}{1.12}
\begin{tabular}{lrrr}
\toprule
\textbf{Profile}
& \textbf{Labeled $n$}
& \textbf{Guided}
& \textbf{Dependent} \\
\midrule
\multicolumn{4}{l}{\textit{Window 1: Before the First Query}} \\
\midrule
Cold Start
& 30 & 3.3\% & \textbf{96.7\%} \\
Oriented
& 97 & \textbf{69.1\%} & 30.9\% \\
Struggling
& 21 & \textbf{71.4\%} & 28.6\% \\
\midrule
\multicolumn{4}{l}{\textit{Window 2: Between Queries}} \\
\midrule
Passive
& 318 & 25.8\% & \textbf{74.2\%} \\
Iterating
& 290 & 32.4\% & \textbf{67.6\%} \\
Debugging
& 56 & 35.7\% & \textbf{64.3\%} \\
Spinning
& 22 & 36.4\% & \textbf{63.6\%} \\
\bottomrule
\end{tabular}
\vspace{-8pt}
\end{table}

\subsection{Demonstration of Downstream Utility: Prediction Tasks}
\label{sec:prediction}

The preliminary evaluation demonstrates how behavioral profiles
can support reactive adaptation during an AI interaction. We
next examine whether the telemetry and behavioral abstractions
provided by \sys{} support two additional predictive decisions:
\textit{query imminence}, whether a learner will submit an AI
query within the next 60 seconds, and \textit{help-seeking
type}, whether an upcoming query reflects guided or dependent
help-seeking. Guided queries articulate a specific need,
question, or concern, whereas dependent queries offload
identifying the problem or determining the next step to the AI.

Together, predicting when a learner is likely to query and
whether that query will be dependent could allow systems to
intervene preemptively and encourage behaviors associated with
more guided help-seeking. For example, 96.7\% of Cold Start
queries were dependent, compared with 30.9\% and 28.6\% for
Oriented and Struggling learners, respectively
(Table~\ref{tab:profile-query}).

\subsubsection{Prediction Tasks}

For \textbf{query-imminence prediction}, the system maintains a
30-second observation window that advances in 5-second steps.
Each window is labeled according to whether the learner submits
an AI query within the following 60 seconds.

For \textbf{help-seeking-type prediction}, each query is paired
with the learner's activity during the 15 seconds before
submission. The model predicts whether the query is guided,
articulating a specific goal, concept, problem, or
misunderstanding, or dependent, offloading identification of
the problem or next step through a vague request, completion
directive, or code without a specific question. This
distinction parallels adaptive and unproductive help-seeking in
prior tutoring research
\cite{aleven2016help,marwan2020unproductive}.

We use GPT-4o to label all queries offline as guided or
dependent. For each query, the model receives the query text,
the learner's code at submission, and the preceding chat
history. The full labeling prompt is provided in
Appendix~\ref{app:query-label-prompt}. Two human raters
independently classified 97 queries and achieved substantial
agreement ($\kappa{=}.897$)~\cite{landis1977measurement}.
Agreement between GPT-4o and the two raters was
$\kappa{=}.709$ and $\kappa{=}.690$, respectively.

For the prediction tasks, semantic labeling inputs were excluded. Both
tasks omit query-composition events
(\texttt{CHAT\_TYPE}, \texttt{CHAT\_DELETE},
\texttt{CHAT\_PASTE}, and \texttt{CHAT\_QUERY}), query text,
source-code content, chat history, and the subsequent AI response. This prevents
query-imminence models from detecting query composition, while
help-seeking-type windows end immediately before submission.
The models therefore rely only on preceding behavioral
telemetry to predict when help-seeking will occur and what form
it will take.

\subsubsection{Feature Representations}

For both tasks, we compare three nested feature representations
to evaluate whether the higher-level abstractions produced by
\sys{} provide predictive value beyond raw telemetry. Features
are computed strictly within each task's observation window:
30 seconds for query imminence and 15 seconds for help-seeking
type.

\begin{itemize}[noitemsep, topsep=3pt]
    \item \textbf{Raw telemetry}: counts of each telemetry
    event type within the task-specific window.

    \item \textbf{+Observable metrics}: adds the observable
    metrics of Section~\ref{sec:quantification}, computed
    within the same window.

    \item \textbf{+Behavioral sequences}: adds features derived
    from the auto-classified behavioral sequences: time in each
    behavior, the current and preceding behavior, and counts of
    transitions between behaviors.
\end{itemize}

\subsubsection{Experimental Protocol}

For each task and feature representation, we train a Random
Forest classifier using Deployment~1 (Instructor A, morning,
$n{=}190$) and evaluate it on held-out Deployment~2
(Instructor A, afternoon, $n{=}113$; Table~\ref{tab:deployment}). Deployment~2 instances
are excluded from model fitting and hyperparameter selection.

We report area under the receiver operating characteristic
curve (AUROC)~\cite{fawcett2006introduction} as the primary
measure of predictive discrimination. AUROC measures how
consistently a model ranks positive instances above negative
instances across classification thresholds.

\subsubsection{Profiles and Help-Seeking Type}

Table~\ref{tab:profile-query} reports guided and dependent
help-seeking across behavioral profiles. In Window~1, 96.7\% of
labeled first queries from Cold Start learners were dependent.
In contrast, first queries from Oriented and Struggling learners
were more often guided (69.1\% and 71.4\%, respectively).
Learners who queried without first editing or executing code
therefore exhibited a markedly different help-seeking
distribution from those who engaged in observable programming
activity before their first query.

Differences were less pronounced in Window~2. Queries following
all four inter-query profiles were more often dependent than
guided, with rates ranging from 63.6\% for Spinning to 74.2\%
for Passive. Because learners may contribute multiple Window~2
observations, these percentages describe profile--query pairs
rather than independent learner groups. We therefore interpret
them as descriptive associations between the activity preceding
a query and the form of help-seeking that follows.

\subsubsection{Prediction Results}
\label{sec:predpower}

Table~\ref{tab:ablation} reports held-out AUROC across the three
nested feature representations. For query-imminence prediction,
raw telemetry achieved an AUROC of 0.689, while the full representation including behavioral-sequence features achieved
0.719. Observable metrics alone performed best at 0.726, an absolute improvement of 0.037 over raw telemetry.

A similar pattern emerged for help-seeking-type prediction. Raw telemetry achieved an AUROC of 0.690, while the full representation including behavioral-sequence features achieved
0.705. Observable metrics alone performed best at 0.717, an absolute improvement of 0.027 over raw telemetry.

Across both tasks, observable metrics provided the largest gain
over raw event counts. Behavioral-sequence features remained
above raw telemetry but did not improve on observable metrics in
this Random Forest~\cite{breiman2001random} evaluation. Together,
these results suggest that window-level summaries of learner
activity capture useful signal about both the timing and form of
help-seeking; Appendix~\ref{app:breakdowns} reports
window-size sensitivity, complete feature contributions, and
profile distributions.

\section{Discussion}

Our findings motivate a central design question: when an AI
tutor can see that a learner has been consistently passive, anticipate that a
query may be approaching, and estimate that the request will reflect dependent help-seeking, how should it respond?
\sys{} provides a foundation for educational system designers to
build around this question. By making learner behavior observable
and computable in real time, \sys{} enables a new class of support
that guides learners toward the behavioral states where
productive AI use can emerge as a natural byproduct rather than as an explicit goal.

Prior approaches to AI overreliance have emphasized AI literacy~\cite{ma2025not}, restrictions on system use,
and guardrails on model responses~\cite{kazemitabaar2024codeaid,
liffiton2023codehelp, hou2024codetailor, kapoor2026exploring}. Our findings suggest a complementary direction:
supporting the learner's behavioral process. In which, rather than asking only how students should
be taught to use AI, we ask how can adaptive systems support the
behaviors that make productive help-seeking a likely outcome.

\subsection{Design Implications}

\textbf{Interpret a query through the behavior that preceded
it.}
In Window~1, 96.7\% of labeled first queries from Cold Start
learners were dependent, compared with 30.9\% for Oriented and
28.6\% for Struggling learners. Although these associations do
not establish causality, they demonstrate that the query alone
provides an incomplete account of the learner's needs. Tutors
should incorporate evidence of prior editing, execution,
thinking, and error recovery when deciding how to respond.

\textbf{Scaffold the learners behaviors.}
Learners who repeatedly query without editing or executing code
may benefit from an intervention that prompts a concrete action
before further assistance, whereas learners who edit without
testing may benefit from being prompted to run and inspect their
code. In our preliminary comparison, behavior-aware prompting
was associated with a reduction in Passive windows from 50.0\%
to 20.7\% and increased observable activity between queries.
Because this comparison was nonrandomized, it provides initial
evidence rather than a causal estimate.

\textbf{Match support to the learner's demonstrated effort.}
The taxonomy suggests that uniform scaffolding is unlikely to
serve every learner equally. A Cold Start learner may require
orientation toward an initial step, a Passive learner may need
encouragement to act independently, and a learner engaged in
extended unsuccessful effort may require more direct support.
The goal is therefore not simply to restrict assistance, but to
provide the level and form of support appropriate to the
learner's current behavioral context.

\subsection{Broader Implications}

\textbf{Toward behaviorally aware learning systems.}
This work enables a new class of behaviorally aware systems that
respond not only to what learners say, but also to how they have
worked leading up to a help-seeking moment. By incorporating this
process-level context, such systems can create conditions for
learners to succeed by preserving productive struggle, supporting
learner agency, and providing scaffolding appropriate to their
recent behavior.

\section{Limitations and Future Work}
\label{sec:limitations}

Our data were collected during short introductory programming
tasks at a single institution, limiting generalizability across
courses, tasks, populations, and learning environments. The
preliminary evaluation compared a morning baseline deployment
with an afternoon intervention deployment without random
assignment. Cohort composition, time of day, or other unmeasured
factors may therefore contribute to the observed behavioral
differences. Future work should evaluate the taxonomy and
interventions across institutions using randomized studies.

The Window~1 and Window~2 profiles were selected through an
exploratory procedure partly informed by task-completion
differences. Their completion rates should therefore be
interpreted as descriptive characteristics rather than
independent validation. Guided and dependent help-seeking
labels were generated by GPT-4o and validated against two human
raters on a subset of queries. Despite substantial agreement,
these labels remain approximations rather than direct
measurements of learners' cognitive engagement.

Our system also uses a single tutor model, GPT-4o. Behavioral
patterns and intervention effects may differ across models with
different capabilities, response styles, or scaffolding
strategies. Future work should test whether the profiles and
prediction tasks remain stable across models.

Finally, the rule-based auto-segmentation pipeline lacks
semantic understanding of learner code and intent. For example,
it may classify edits as Debugging while an unresolved error
remains even when the learner is implementing an unrelated
feature. Future work should combine behavioral telemetry with
lightweight semantic code analysis and test whether reduced
Passive re-querying improves knowledge retention, transfer, and
long-term help-seeking behavior.

\section{Conclusion}

We presented \sys{}, a dataset and real-time behavioral
abstraction pipeline that makes the context surrounding
learner--AI interactions computable. Across 480 learners in four
introductory Python deployments, \sys{} captures approximately
180K telemetry events, 13,633 auto-classified behavioral
segments, and 27 observable metrics validated against expert
labels and student self-reports.

We derived a three-window taxonomy characterizing activity
before the first AI query, between consecutive queries, and
across the session. In a preliminary between-deployment
evaluation, behavior-aware prompting was associated with a
reduction in Passive inter-query windows from 50.0\% to 20.7\%.
Observable metrics also improved held-out AUROC from 0.689 to
0.726 for query imminence and from 0.690 to 0.717 for
help-seeking type.

We release the \sys{} dataset, taxonomy, classifier, and code
through the repository linked in the Introduction to support AI
tutors that respond not only to what learners say, but also to
what they do.

\bibliographystyle{ACM-Reference-Format}
\bibliography{sample-base}

\clearpage
\onecolumn
\raggedbottom

\appendix
 
\section{Behavioral Labeling Interface}
\label{app:labeling-interface}
 
\begin{figure}[H]
\centering
\includegraphics[width=0.85\textwidth]{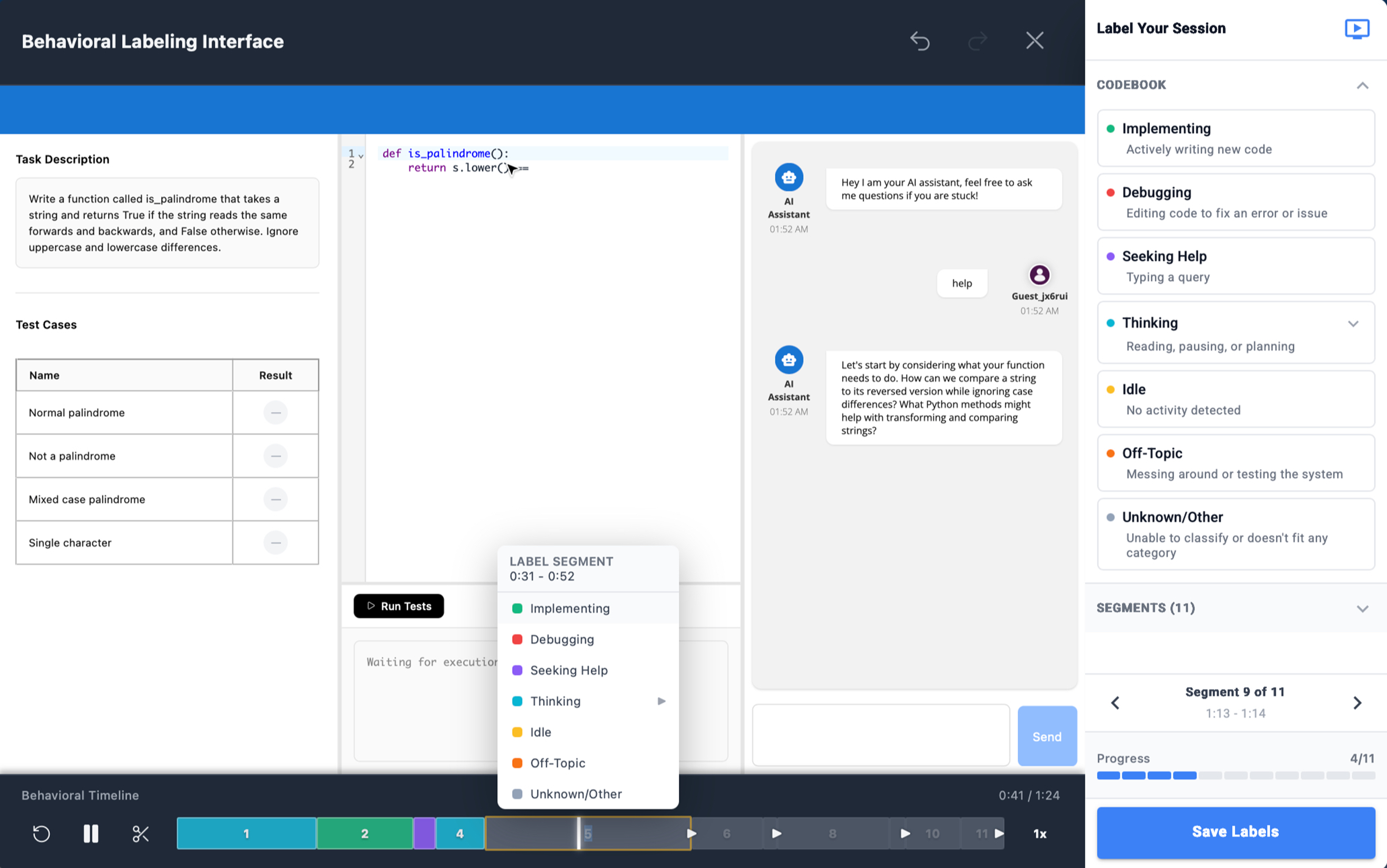}
\caption{The Behavioral Labeling Interface used by expert labelers to annotate student sessions. The interface presents a session replay, a timeline for placing segment boundaries, and the behavioral codebook for classification.}
\Description{Screenshot of the labeling tool with a session replay, an annotatable timeline, and the behavioral codebook.}
\label{fig:labeling-interface}
\end{figure}

\section{Codebook Change History}
\label{app:codebook-history}
 
The behavioral codebook was iteratively refined across 10 pilot labeling sessions through weekly reconciliation meetings. Table~\ref{tab:codebook-history} documents each revision with the corresponding date and specific change.
 
\begin{table}[H]
\centering
\caption{Codebook revision history across the pilot labeling phase, derived from platform configuration changes.}
\label{tab:codebook-history}
\small
\renewcommand{\arraystretch}{1.2}
\begin{tabular}{lp{5.8cm}}
\toprule
\textbf{Date} & \textbf{Change} \\
\midrule
Feb 2 & Initial codebook: Implementing, Debugging, Thinking (with subtypes: Task, Code, Error, LLM Response), Idle, Off-Topic, Unknown. \\
Feb 3 & Added \textit{Seeking Help} (``Typing a query or reading LLM response''). \\
Feb 9 & Revised \textit{Seeking Help} description to ``Typing a query'' only, removing ``reading LLM response'' to resolve overlap with Thinking: LLM Response subtype. \\
Feb 10 & Added \textit{Testing} (``Running code and reviewing results''). \\
Mar 11 & Removed \textit{Testing} from manual labeling codebook (captured by auto-segmentation rules). Added \textit{Thinking: Pre-query} subtype (``Asking LLM'') to capture formulation pauses before chat queries. \\
Mar 11 & Codebook finalized. All labelers confirmed agreement on category definitions and segmentation rules. \\
\bottomrule
\end{tabular}
\end{table}
 
\clearpage
\section{Telemetry Event Schema}
\label{app:schema}
 
\begin{table}[H]
\centering
\caption{Full telemetry event schema with example payloads from live deployments. Each event includes a millisecond timestamp, source region, and event-specific payload. Long event names and payloads are abbreviated for space.}
\label{tab:telemetry-full}
\footnotesize
\setlength{\tabcolsep}{2.5pt}
\renewcommand{\arraystretch}{1.02}
\begin{tabular}{ll>{\raggedright\arraybackslash}p{6.8cm}}
\toprule
\textbf{Event Type} & \textbf{Description} & \textbf{Example Payload} \\
\midrule
\multicolumn{3}{l}{\textit{Code Editor (11 types)}} \\
\midrule
\texttt{CODE\_TYPE} & Typed characters & \texttt{\{code:"l", changes:[\{to:0, from:0, text:"l"\}], raw\_action:"input.type"\}} \\
\texttt{CODE\_DELETE} & Backspace/Delete & \texttt{\{code:"...grade\_boo", changes:[\{to:101, from:100, text:""\}], raw\_action:"delete.backward"\}} \\
\texttt{CODE\_DELETE\_SEL.} & Deleted selection & \texttt{\{code:"", changes:[\{to:145, from:0, text:""\}], raw\_action:"delete.selection"\}} \\
\texttt{CODE\_PASTE} & Pasted content & \texttt{\{code:"...append([78,84,91])", changes:[\{to:75, from:75, text:"[78,84,91]"\}], raw\_action:"input.paste"\}} \\
\texttt{CODE\_COPY} & Copied to clipboard & \texttt{\{selected\_text:"grade\_book = [[88,92,75],...]"\}} \\
\texttt{CODE\_CUT} & Cut to clipboard & \texttt{\{code:"", changes:[\{to:186, from:0, text:""\}], raw\_action:"delete.cut"\}} \\
\texttt{CODE\_UNDO} & Ctrl+Z & \texttt{\{code:"...append()", changes:[\{to:88, from:76, text:""\}], raw\_action:"undo"\}} \\
\texttt{CODE\_REDO} & Ctrl+Y & \texttt{\{changes:[...], raw\_action:"redo"\}} \\
\texttt{CODE\_SELECT} & Highlighted text & \texttt{\{selected\_text:"print(grade\_book)"\}} \\
\texttt{CODE\_INDENT} & Tab key & \texttt{\{code:"...for grade in grade\_book:{\textbackslash}n~~~~", changes:[\{to:96, from:96, text:"~~~~"\}], raw\_action:"input.indent"\}} \\
\texttt{CODE\_UNKNOWN} & Unrecognized edit & \texttt{\{code:"...li", changes:[\{to:57, from:57, text:"i"\}], raw\_action:"input.type.compose"\}} \\
\midrule
\multicolumn{3}{l}{\textit{Terminal (6 types)}} \\
\midrule
\texttt{TERMINAL\_RUN} & Clicked Run Tests & \texttt{\{code\_length:180\}} \\
\texttt{TERMINAL\_OUTPUT} & stdout printed & \texttt{\{output:"95"\}} \\
\texttt{TEST\_CASE\_RESULT} & Test results & \texttt{\{passed\_count:2, total\_tests:3, passrate:66.7, detailed\_results:[\{passed:true, test\_name:"Test Case 1"\}, ...]\}} \\
\texttt{TERMINAL\_ERROR} & Runtime error & \texttt{\{type:"UnknownError", message:"Traceback...IndexError: list assignment index out of range"\}} \\
\texttt{TERMINAL\_SELECT} & Highlighted output & \texttt{\{selected\_text:"IndexError: list assignment index out of range"\}} \\
\texttt{TERMINAL\_COPY} & Copied output & \texttt{\{selected\_text:"IndexError: list assignment index out of range"\}} \\
\midrule
\multicolumn{3}{l}{\textit{Chat (9 types)}} \\
\midrule
\texttt{CHAT\_TYPE} & Typed in chat & \texttt{\{text:"h"\}} \\
\texttt{CHAT\_DELETE} & Deleted in chat & \texttt{\{key\_used:"Backspace", text\_before:"...command fr", text\_after:"...command f", cursor\_position:34\}} \\
\texttt{CHAT\_PASTE} & Pasted into chat & \texttt{\{pasted\_text:"IndexError: list assignment index out of range"\}} \\
\texttt{CHAT\_QUERY} & Sent message to AI & \texttt{\{text:"hi", length:2\}} \\
\texttt{CHAT\_RESPONSE} & AI response & \texttt{\{content:"Great start! Let's tackle this step by step...", length:322, latency\_ms:2495\}} \\
\texttt{CHAT\_SELECT\_INPUT} & Highlighted input & \texttt{\{selected\_text:"what would code be"\}} \\
\texttt{CHAT\_SELECT\_HIST.} & Highlighted history & \texttt{\{selected\_text:"what should I do?"\}} \\
\texttt{CHAT\_COPY\_INPUT} & Copied from input & \texttt{\{selected\_text:"..."\}} \\
\texttt{CHAT\_COPY\_HIST.} & Copied from history & \texttt{\{selected\_text:"what should I do?"\}} \\
\midrule
\multicolumn{3}{l}{\textit{Task / Tests (4 types)}} \\
\midrule
\texttt{TASK\_SELECT} & Highlighted task & \texttt{\{selected\_text:"[78, 84, 91"\}} \\
\texttt{TASK\_COPY} & Copied from task & \texttt{\{selected\_text:"[78, 84, 91]"\}} \\
\texttt{TEST\_SELECT} & Highlighted test & \texttt{\{selected\_text:"New"\}} \\
\texttt{TEST\_COPY} & Copied from test & \texttt{\{selected\_text:"Print all scores"\}} \\
\midrule
\multicolumn{3}{l}{\textit{Global (5 types)}} \\
\midrule
\texttt{MOUSE\_CLICK} & Clicked in workspace & \texttt{\{x:0.588, y:0.183, region:"CODE\_EDITOR", tag:"DIV", className:"cm-content"\}} \\
\texttt{MOUSE\_MOVE} & Mouse position (50ms) & \texttt{\{x:0.384, y:0.272, region:"CODE\_EDITOR"\}} \\
\texttt{TAB\_STATE} & Browser tab switch & \texttt{\{visible:false\}} \\
\texttt{WINDOW\_RESIZE} & Browser resized & \texttt{\{width:1588, height:901\}} \\
\texttt{PANEL\_RESIZE} & Panel resized & \texttt{\{panel:"left", newSize:456\}} \\
\midrule
\multicolumn{3}{l}{\textit{Session (2 types)}} \\
\midrule
\texttt{SESSION\_START} & Entered workspace & \texttt{\{window\_width:1588, window\_height:901\}} \\
\texttt{SESSION\_END} & Left workspace & \texttt{\{duration\_ms:542000\}} \\
\bottomrule
\end{tabular}
\end{table}
 
\clearpage
\section{LLM System Prompt}
\label{app:prompt}
 
The system prompt provided to GPT-4o for every student
interaction is shown in Figure~\ref{fig:sysprompt}. The prompt
implements the three-level scaffolding framework described in
Section~\ref{sec:llm-integration}. The behavioral context blocks
referenced in its final paragraph are reproduced in
Appendix~\ref{app:profiles}. These blocks could override the
default escalation order by directing the model to begin at a
higher scaffolding level, but did not override the prohibition
against providing runnable code.
 
\begin{figure}[H]
\begin{wideprompt}
\begin{multicols}{2}\small
\textbf{Role:} You are a supportive programming tutor embedded in an educational coding environment. Your role is to help students develop problem-solving skills through guided inquiry, not to solve problems for them.
 
\textbf{Task Context:} You will be provided with the task description, the student's current code, their test results, and the conversation history. Use the conversation history to track escalation level and whether the student has made progress.
 
\textbf{Core Approach --- Three-Level Scaffolding:}
 
\textit{Level 1 --- Socratic Questioning (default):} Ask one targeted question that directs the student's attention to the relevant part of their code or problem. Do not provide hints at this level.
 
\textit{Level 2 --- Conceptual Hint:} Name the relevant concept or method without showing syntax, then ask a follow-up question.
 
\textit{Level 3 --- Concrete Scaffolding:} Provide a pseudocode skeleton with blanks for the current step only. Never show the structure of the full solution. All specific values, indices, and strings must be replaced with \texttt{\_\_\_}.
 
\textbf{Absolute Rules:}
\begin{enumerate}[noitemsep]
\item Never write runnable code in any programming language. When showing patterns, use only pseudocode with blank placeholders (\texttt{\_\_\_}).
\item Never scaffold more than one step at a time.
\item Never skip escalation levels. Escalate exactly one level at a time.
\item When the student moves to a new step or concept, always reset to Level 1.
\end{enumerate}
 
\textbf{Escalation Rules:}
Start at Level 1 for each new concept. After each response, assess whether the student made progress. If the current level did not result in progress, escalate one level. Only one escalation per response. If the student has no code and expresses uncertainty, begin at Level 2 with a conceptual hint about the first step only. When the student completes a step and moves to a new concept, reset to Level 1.
 
\textbf{Response Guidelines:}
Keep responses concise (2--4 sentences plus one question). Ask only one question per response. Reference specific lines, variables, or test failures from the student's actual code.
 
\textbf{Behavioral Context Override:}
If a behavioral context block is present, it describes what the student was doing before they asked. Use it to adjust scaffolding approach and escalation pace. The behavioral context never overrides the rule against writing runnable code; it only changes how scaffolding is delivered (e.g., starting at a higher level, trying a different angle, acknowledging effort).
\end{multicols}
\end{wideprompt}
\caption{The LLM tutoring system prompt provided to GPT-4o for every student interaction, implementing three escalating levels of scaffolding with a behavioral context override.}
\Description{The complete tutoring system prompt, laid out in two columns.
It defines the tutor role, the task context provided at each interaction,
and three escalating scaffolding levels: Socratic questioning, conceptual
hint, and concrete scaffolding with pseudocode blanks. It then lists
absolute rules forbidding runnable code and multi-step scaffolds,
escalation rules for moving one level at a time and resetting on new
concepts, response guidelines limiting length and questions, and a
behavioral context override explaining how appended behavior blocks adjust
scaffolding without overriding the code prohibition.}
\label{fig:sysprompt}
\end{figure}

\section{Foundation Dataset Structure}
\label{app:dataset}
 
Table~\ref{tab:dataset-layers} summarizes the abstraction layers of the released foundation dataset and its overall scale.
 
\begin{table}[H]
\centering
\caption{Foundation dataset structure. Layers A--C
comprise the \sys{} foundation; Layer D is derived in the \sys{} Taxonomy.}
\label{tab:dataset-layers}
\small
\begin{tabular}{llr}
\toprule
\textbf{Layer} & \textbf{Representation} & \textbf{Scale} \\
\midrule
A: Raw Telemetry        & Timestamped IDE events       & ${\sim}$180K \\
B: Observable Metrics   & Effort intensity per window   & 27 features \\
C: Behavioral Sequences & Auto-classified segments      & 13,633 \\
D: Behavioral Profiles    & Clustered profiles            & 10 profiles \\
\midrule
Student sessions        &                               & 480 \\
AI interactions         &                               & 1,386 \\
Queries                 &                               & 840 \\
\bottomrule
\end{tabular}
\end{table}
 
\clearpage
\section{Auto-Segmentation Algorithm}
\label{app:algorithm}
 
Algorithm~\ref{alg:autoseg} presents the seven-step
auto-segmentation pipeline that translates raw telemetry events
into labeled behavioral sequences. Each step corresponds to a
rule derived from expert-labeler consensus during codebook
development (Section~\ref{sec:codebook-development}). 
 
\begin{algorithm}[H]
\caption{Auto-Behavioral Classification Pipeline}
\label{alg:autoseg}
\footnotesize
\begin{algorithmic}[1]
 
\Require Events $E = [e_1, \ldots, e_n]$ with timestamps relative to session start; session duration $D$
\Ensure Ordered list of behavioral segments $S$, each with behavior label and thinking subtype
 
\Statex \textbf{Step 1: Build Major Segments}
\State Classify each event by category: \textsc{Code}, \textsc{Terminal}, \textsc{ChatInput}, \textsc{ChatResponse}
\State Group consecutive same-category events into segments
\State \textbf{if} two consecutive \textsc{Code} events are separated by $\geq 6$s, split into two segments
\State Assign initial behavior: \textsc{Code} $\to$ Implementing, \textsc{Terminal} $\to$ Testing, \textsc{ChatInput} $\to$ Seeking Help, \textsc{ChatResponse} $\to$ Thinking
 
\Statex \textbf{Step 2: Fill Gaps with Thinking}
\For{each gap between consecutive segments}
  \If{gap $\geq 3$s}
    \State Insert Thinking segment spanning the gap
  \Else
    \State Extend the preceding segment to close the gap
  \EndIf
\EndFor
\State Apply same logic to gaps before the first segment and after the last segment
 
\Statex \textbf{Step 3: Merge Short Testing Segments}
\For{each Testing segment with duration $< 1.5$s}
  \State Absorb into the adjacent segment (prefer next; fall back to previous)
\EndFor
 
\Statex \textbf{Step 4: Absorb Pre-Query Pauses}
\For{each Thinking segment of duration $\leq 5$s}
  \If{preceded by Implementing or Debugging \textbf{and} followed by Seeking Help}
    \State Store pause duration as metadata on the Seeking Help segment
    \State Extend the preceding segment to cover the pause
    \State Remove the Thinking segment
  \EndIf
\EndFor
 
\Statex \textbf{Step 5: Apply Error State}
\State Maintain flag $\textit{hasUnresolvedError} \gets \textsc{false}$
\For{each segment in chronological order}
  \If{segment contains \texttt{TERMINAL\_ERROR} or a failed \texttt{TEST\_CASE\_RESULT}}
    \State $\textit{hasUnresolvedError} \gets \textsc{true}$
  \EndIf
  \If{segment contains a \texttt{TEST\_CASE\_RESULT} where all tests pass}
    \State $\textit{hasUnresolvedError} \gets \textsc{false}$
  \EndIf
  \If{segment is Implementing \textbf{and} $\textit{hasUnresolvedError}$}
    \State Relabel segment as Debugging
  \EndIf
\EndFor
 
\Statex \textbf{Step 6: Classify Thinking Subtypes}
\For{each Thinking segment in chronological order}
  \If{no code, terminal, or chat activity has occurred yet}
    \State Label as \textit{Thinking: Task} \Comment{Reading the task description}
  \ElsIf{preceding segment was Seeking Help}
    \State Label as \textit{Thinking: Response} \Comment{Reading AI reply}
  \ElsIf{most recent terminal run produced an unresolved error}
    \State Label as \textit{Thinking: Error} \Comment{Reading an error message}
  \Else
    \State Label as \textit{Thinking: Code} \Comment{Reviewing own code}
  \EndIf
\EndFor
 
\Statex \textbf{Step 7: Post-Process}
\State Fix any remaining unlabeled segments (terminal events $\to$ Testing; otherwise $\to$ Thinking)
\State Merge consecutive segments with the same behavior label
\State Re-index all segment IDs
 
\State \Return $S$
\end{algorithmic}
\end{algorithm}
 
\clearpage
\section{Behavior-Aware Intervention Prompts}
\label{app:profiles}
 
In the treatment condition
(Section~\ref{sec:prelim}), the system appended one of five
behavior-aware context blocks to the student's query. Cold Start
was used for a learner's first query when no code edit or
terminal run had occurred. For subsequent queries, the system
selected among Passive, Iterating, Debugging, and Spinning using
online activity rules derived during system development. These
deployment-time rules used the same profile names as the later
taxonomy but represented precursor heuristic states rather than
assignments from the revised clustering procedure. The prompts
below reproduce the exact intervention text used during
deployment. The revised taxonomy was subsequently applied to
both deployments to compare their Window~2 profile
distributions.
 
\begin{promptbox}{Cold Start}
This is the student's first time asking for help. They have not written or run any code yet. Give them something concrete to start with --- a pseudocode scaffold with blanks to fill in. End your response with a single sentence starting with ``Try:'' that gives them that pseudocode starting point so they know what to write next.
\end{promptbox}
 
\begin{promptbox}{Passive}
This student received help but has not written or run any code since. Try a different angle from your last response --- do not repeat the same explanation. Do NOT escalate your scaffolding level --- stay at the same level but approach it differently. If the student asks ``how to fix'' or requests the answer directly, acknowledge, redirect, and re-ask with pseudocode. If the student has just completed a step and is asking what to do next, start at Level 1 for the new step. If the question is purely conceptual, answer it naturally. Otherwise, end your response with a single sentence starting with ``Try:'' that gives them one concrete next action using pseudocode with blanks if code is involved.
\end{promptbox}
 
\begin{promptbox}{Iterating}
This student wrote some code but has not tested it. Do not give more explanation. If their current code would produce meaningful output (e.g.\ contains a print statement, an expression, or enough logic to show a result), end your response with a single sentence starting with ``Try:'' telling them to run it. If running it would produce no output, end with a ``Try:'' that asks them to add a specific print statement first so they can see what their code is doing.
\end{promptbox}
 
\begin{promptbox}{Debugging}
This student has been coding, testing, and hitting errors. They are engaged and struggling. Recognize their effort. Name the specific error and the misconception behind it so they can read errors independently. Escalate to Level 2 or 3 immediately. End your response with a single sentence starting with ``Try:'' giving them the targeted fix as a pseudocode pattern with blanks.
\end{promptbox}
 
\begin{promptbox}{Spinning}
This student has run their code many times and keeps hitting errors. They do not need encouragement --- they need to be unblocked. Name the concept or method they need, explain briefly why their current approach fails, and escalate to Level 3 immediately. End your response with a single sentence starting with ``Try:'' that gives them the precise pseudocode pattern they need, with blanks for the specific values.
\end{promptbox}
 
\clearpage
\section{Help-Seeking Type Labeling Prompt}
\label{app:query-label-prompt}
 
The system prompt shown in Figure~\ref{fig:labelprompt} was provided to GPT-4o to classify
each student query as guided or dependent. The model received
the student's query, current code, and preceding chat history.
These semantic inputs were used only to generate the offline
help-seeking labels and were excluded from the prediction
models described in Section~\ref{sec:prediction}.
 
\begin{figure}[H]
\begin{wideprompt}
\begin{multicols}{2}\small
 
You are an educational assistant analyzing student help
requests in introductory programming courses.
 
Given a student query, their current code, and chat history,
classify the query as either \texttt{GUIDED} or
\texttt{DEPENDENT}.
 
\textbf{GUIDED ---} The student demonstrates independent
thinking. They have identified what they need help with and are
actively steering their learning. This includes:
 
\begin{itemize}[noitemsep]
    \item Asking a specific question about a concept
    (``What is a nested loop?'')
    \item Identifying a specific problem or confusion
    (``I'm not sure what to put in the print statement'')
    \item Describing what they tried and what went wrong
    (``I tried using a for loop but it only prints the first
    item'')
    \item Asking how to approach a specific step
    (``How do I iterate through each student's grades?'')
    \item Requesting clarification on a specific point from a
    prior AI response
    (``What do you mean by iterating over the inner list?'')
    \item Answering the AI's question with specific information
    (``The error is IndexError on line 5'')
\end{itemize}
 
The key indicator: the student has done some cognitive work to
formulate what they need. The query communicates a specific
need, question, or confusion, even if brief.
 
\textbf{DEPENDENT ---} The student is offloading cognitive work
to the AI with minimal independent effort. This includes:
 
\begin{itemize}[noitemsep]
    \item Pasting code with no question or description of the
    problem (implicit ``fix this for me'')
    \item Vague requests with no specifics
    (``help,'' ``it doesn't work,'' ``idk'')
    \item Pure acknowledgments that delegate next steps
    (``ok do that,'' ``yeah,'' ``sure,'' ``go ahead'')
    \item Empty or near-empty messages
    (``?,'' stray characters)
    \item Requests that ask the AI to do the work
    (``can you just write it for me,'' ``give me the code'')
    \item Repeating the assignment prompt or pasting
    instructions without any attempt or question
    \item Answering the AI's question with no effort
    (``idk,'' ``I don't know,'' ``you tell me'')
\end{itemize}
 
The key indicator: the student has not done cognitive work to
identify what they need. They are asking the AI to do the
thinking for them.
 
\textbf{Boundary cases --- apply these rules:}
 
\begin{itemize}[noitemsep]
    \item ``I don't know how to do X'' $\rightarrow$
    \texttt{GUIDED} (they identified what they do not know)
    \item ``I don't know'' or ``I'm stuck'' alone, with no
    specifics $\rightarrow$ \texttt{DEPENDENT}
    \item ``Is this right?'' with code $\rightarrow$
    \texttt{DEPENDENT}
    \item ``Is this right? I'm not sure if my loop handles the
    last element'' $\rightarrow$ \texttt{GUIDED}
    \item Code pasted with ``Why does this print the whole list
    instead of individual scores?'' $\rightarrow$
    \texttt{GUIDED}
    \item Code pasted with no question $\rightarrow$
    \texttt{DEPENDENT}
    \item ``What about edge cases?'' in the context of an
    ongoing conversation $\rightarrow$ \texttt{GUIDED}
    \item ``Ok,'' ``Thanks,'' or ``Got it'' alone
    $\rightarrow$ \texttt{DEPENDENT}
    \item ``Ok, but how does that work with nested lists?''
    $\rightarrow$ \texttt{GUIDED}
\end{itemize}
 
\textbf{Confidence:}
 
\begin{itemize}[noitemsep]
    \item \texttt{high}: clearly guided or clearly dependent
    \item \texttt{medium}: leans one way but has some ambiguity
    \item \texttt{low}: genuinely on the boundary
\end{itemize}
 
Return only a JSON object:
 
\medskip
\noindent\texttt{\{}\\
\texttt{~~"queryEngagement": "guided" or "dependent",}\\
\texttt{~~"rationale": "1 sentence explaining why",}\\
\texttt{~~"confidence": "high/medium/low"}\\
\texttt{\}}
 
\end{multicols}
\end{wideprompt}
\caption{The help-seeking type labeling prompt provided to GPT-4o to classify each student query as guided or dependent from the query text, the student's code at submission time, and the preceding chat history.}
\Description{The complete help-seeking classification prompt, laid out in
two columns. It instructs the model to label each student query as guided
or dependent, defines guided queries as demonstrating independent thinking
with bulleted examples, defines dependent queries as offloading cognitive
work with bulleted examples, lists explicit boundary-case rules mapping
ambiguous phrasings to labels, defines high, medium, and low confidence,
and specifies a JSON output object with the label, a one-sentence
rationale, and a confidence level.}
\label{fig:labelprompt}
\end{figure}
 
\clearpage
\section{Task Descriptions}
\label{app:tasks}
 
Full task descriptions and test cases are reproduced below as presented to students.
 
\medskip
\noindent Students were given a pre-populated list of song names and asked to complete four list manipulation steps within 15 minutes.
 
\begin{promptbox}{Deployments 1--2: Playlist (Python, 15 min)}
You are given a list of song names called \texttt{playlist}. Complete the following steps:
\begin{enumerate}[noitemsep]
\item The song at index 4 was added by mistake. Replace it with ``Purple Rain''.
\item Remove the last 3 songs from the playlist using slicing.
\item Reverse the playlist in place.
\item Print every other song in the playlist using slicing with a step value.
\end{enumerate}
\smallskip
\textbf{Test cases:}
\begin{enumerate}[noitemsep, label=\arabic*.]
\item \texttt{"Purple Rain" in playlist} $\to$ \texttt{True}
\item \texttt{len(playlist)} $\to$ \texttt{5}
\item \texttt{playlist[0]} $\to$ \texttt{"Purple Rain"}
\item \texttt{playlist[::2]} $\to$ \texttt{["Purple Rain", "Imagine", "Bohemian Rhapsody"]}
\end{enumerate}
\end{promptbox}
 
\medskip
\noindent Students were given a nested list of test scores and asked to complete three list manipulation steps within 10 minutes.
 
\begin{promptbox}{Deployments 3--4: Grade Book (Python, 10 min)}
You are given a list of lists called \texttt{grade\_book}, where each inner list contains a student's test scores. Complete the following steps:
\begin{enumerate}[noitemsep]
\item A new student joined the class with scores \texttt{[78, 84, 91]}. Add their scores to \texttt{grade\_book}.
\item The first student's third score was entered incorrectly. Update it to \texttt{80}.
\item Using a nested for loop, print every individual score in \texttt{grade\_book}, one per line.
\end{enumerate}
\smallskip
\textbf{Test cases:}
\begin{enumerate}[noitemsep, label=\arabic*.]
\item \texttt{len(grade\_book)} $\to$ \texttt{4}
\item \texttt{grade\_book[3]} $\to$ \texttt{[78, 84, 91]}
\item \texttt{grade\_book[0][2]} $\to$ \texttt{80}
\item Nested loop prints all 12 scores, one per line.
\end{enumerate}
\end{promptbox}

\clearpage
\section{Prediction Breakdowns: Observation Windows, Feature Contributions, and Profiles}
\label{app:breakdowns}
 
 
\begin{table}[H]
\centering
\caption{Held-out AUROC across observation-window sizes for both
prediction tasks and all three feature layers. The 30-second
query-imminence row and 15-second help-seeking-type row correspond to
Table~\ref{tab:ablation}; bold marks the configuration reported
in the subsequent tables.}
\label{tab:window-sweep}
\small
\setlength{\tabcolsep}{6pt}
\renewcommand{\arraystretch}{1.12}
\begin{tabular}{llrrccc}
\toprule
\textbf{Task} & \textbf{Window} & \textbf{Test $n$} & \textbf{Positive} & \textbf{Raw} & \textbf{+Obs.} & \textbf{+Seq.} \\
\midrule
Query imminence & 15\,s & 11,160 & 37.4\% & 0.657 & 0.690 & 0.707 \\
 & 30\,s & 10,825 & 38.2\% & 0.689 & \textbf{0.726} & 0.719 \\
 & 45\,s & 10,492 & 38.7\% & 0.701 & 0.734 & 0.720 \\
 & 60\,s & 10,165 & 39.1\% & 0.700 & 0.730 & 0.713 \\
\midrule
Help-seeking type & 15\,s & 499 & 33.1\% & 0.690 & \textbf{0.717} & 0.705 \\
 & 30\,s & 499 & 33.1\% & 0.696 & 0.680 & 0.694 \\
 & 45\,s & 498 & 33.1\% & 0.691 & 0.674 & 0.696 \\
 & 60\,s & 495 & 33.3\% & 0.678 & 0.649 & 0.667 \\
\bottomrule
\end{tabular}
\end{table}
 
\begin{table}[H]
\centering
\caption{Complete feature-contribution list (Random-Forest importances, all features with importance $>0$) for query imminence at its reported configuration (30\,s window, observable layer).}
\label{tab:contrib-imminence}
\scriptsize
\setlength{\tabcolsep}{4pt}
\renewcommand{\arraystretch}{1.05}
\begin{tabular}{lr@{\hspace{18pt}}lr}
\toprule
\textbf{Feature} & \textbf{Imp.} & \textbf{Feature} & \textbf{Imp.} \\
\midrule
\metric{metric__time_in_chat_s} & 0.1465 & \metric{metric__error_to_edit_s} & 0.0060 \\
\metric{metric__longest_idle_s} & 0.0761 & \metric{raw__TERMINAL_OUTPUT} & 0.0058 \\
\metric{raw__event_density_per_s} & 0.0575 & \metric{raw__CODE_SELECT} & 0.0057 \\
\metric{metric__time_in_editor_s} & 0.0573 & \metric{metric__failed_test_to_edit_s} & 0.0053 \\
\metric{raw__event_count} & 0.0573 & \metric{raw__TEST_CASE_RESULT} & 0.0043 \\
\metric{raw__MOUSE_MOVE} & 0.0525 & \metric{metric__max_consecutive_errors} & 0.0040 \\
\metric{raw__CHAT_RESPONSE} & 0.0507 & \metric{raw__TERMINAL_ERROR} & 0.0038 \\
\metric{metric__time_in_terminal_s} & 0.0452 & \metric{metric__terminal_errors} & 0.0037 \\
\metric{raw__MOUSE_CLICK} & 0.0435 & \metric{raw__CODE_PASTE} & 0.0028 \\
\metric{metric__thinking_time_s} & 0.0393 & \metric{raw__CODE_UNKNOWN} & 0.0027 \\
\metric{raw__TAB_STATE} & 0.0278 & \metric{raw__CODE_COPY} & 0.0026 \\
\metric{metric__net_code_growth} & 0.0276 & \metric{raw__PANEL_RESIZE} & 0.0026 \\
\metric{metric__chars_inserted} & 0.0263 & \metric{raw__CODE_DELETE_SELECTION} & 0.0020 \\
\metric{metric__time_in_task_s} & 0.0248 & \metric{raw__SESSION_START} & 0.0020 \\
\metric{raw__CODE_TYPE} & 0.0247 & \metric{metric__failed_test_self_fix} & 0.0020 \\
\metric{metric__delete_type_ratio} & 0.0240 & \metric{raw__CODE_INDENT} & 0.0019 \\
\metric{metric__code_edit_rate} & 0.0236 & \metric{metric__error_self_fix} & 0.0017 \\
\metric{metric__code_edits} & 0.0230 & \metric{raw__TASK_COPY} & 0.0016 \\
\metric{metric__chars_deleted} & 0.0218 & \metric{raw__TASK_SELECT} & 0.0014 \\
\metric{metric__time_in_tests_s} & 0.0198 & \metric{raw__CHAT_SELECT_HISTORY} & 0.0013 \\
\metric{raw__CODE_DELETE} & 0.0152 & \metric{raw__CHAT_SELECT_INPUT} & 0.0011 \\
\metric{metric__code_deletes} & 0.0148 & \metric{raw__TERMINAL_SELECT} & 0.0007 \\
\metric{raw__TERMINAL_RUN} & 0.0072 & \metric{raw__CODE_UNDO} & 0.0007 \\
\metric{metric__error_reading_time_s} & 0.0070 & \metric{raw__TERMINAL_COPY} & 0.0003 \\
\metric{metric__terminal_runs} & 0.0069 & \metric{raw__CODE_CUT} & 0.0002 \\
\metric{raw__WINDOW_RESIZE} & 0.0069 & \metric{raw__TEST_SELECT} & 0.0001 \\
\metric{metric__mean_time_between_runs_s} & 0.0062 &  &  \\
\bottomrule
\end{tabular}
\end{table}
 
\begin{table}[H]
\centering
\caption{Complete feature-contribution list (Random-Forest importances, all features with importance $>0$) for help-seeking type at its reported configuration (15\,s window, observable layer).}
\label{tab:contrib-type}
\scriptsize
\setlength{\tabcolsep}{4pt}
\renewcommand{\arraystretch}{1.05}
\begin{tabular}{lr@{\hspace{18pt}}lr}
\toprule
\textbf{Feature} & \textbf{Imp.} & \textbf{Feature} & \textbf{Imp.} \\
\midrule
\metric{metric__longest_idle_s} & 0.1442 & \metric{raw__TERMINAL_OUTPUT} & 0.0064 \\
\metric{metric__time_in_editor_s} & 0.0984 & \metric{metric__max_consecutive_errors} & 0.0054 \\
\metric{raw__event_count} & 0.0922 & \metric{raw__WINDOW_RESIZE} & 0.0050 \\
\metric{raw__event_density_per_s} & 0.0881 & \metric{raw__TERMINAL_ERROR} & 0.0049 \\
\metric{metric__time_in_chat_s} & 0.0868 & \metric{raw__TERMINAL_SELECT} & 0.0048 \\
\metric{raw__MOUSE_MOVE} & 0.0786 & \metric{metric__terminal_errors} & 0.0045 \\
\metric{raw__MOUSE_CLICK} & 0.0625 & \metric{raw__CODE_SELECT} & 0.0042 \\
\metric{metric__time_in_terminal_s} & 0.0501 & \metric{raw__CODE_COPY} & 0.0039 \\
\metric{raw__CHAT_RESPONSE} & 0.0422 & \metric{raw__TERMINAL_COPY} & 0.0036 \\
\metric{metric__thinking_time_s} & 0.0361 & \metric{raw__TASK_COPY} & 0.0025 \\
\metric{metric__time_in_task_s} & 0.0173 & \metric{raw__TASK_SELECT} & 0.0025 \\
\metric{metric__code_edit_rate} & 0.0170 & \metric{raw__TEST_CASE_RESULT} & 0.0024 \\
\metric{metric__code_edits} & 0.0154 & \metric{metric__error_reading_time_s} & 0.0022 \\
\metric{raw__CODE_TYPE} & 0.0137 & \metric{metric__mean_time_between_runs_s} & 0.0010 \\
\metric{metric__net_code_growth} & 0.0132 & \metric{raw__CHAT_SELECT_INPUT} & 0.0009 \\
\metric{metric__chars_inserted} & 0.0123 & \metric{raw__CHAT_SELECT_HISTORY} & 0.0005 \\
\metric{metric__chars_deleted} & 0.0118 & \metric{raw__PANEL_RESIZE} & 0.0004 \\
\metric{metric__code_deletes} & 0.0109 & \metric{raw__CODE_DELETE_SELECTION} & 0.0003 \\
\metric{raw__CODE_DELETE} & 0.0109 & \metric{metric__failed_test_self_fix} & 0.0002 \\
\metric{metric__time_in_tests_s} & 0.0100 & \metric{raw__CODE_PASTE} & 0.0002 \\
\metric{metric__terminal_runs} & 0.0084 & \metric{metric__error_to_edit_s} & 0.0002 \\
\metric{metric__delete_type_ratio} & 0.0083 & \metric{metric__error_self_fix} & 0.0001 \\
\metric{raw__TERMINAL_RUN} & 0.0082 & \metric{metric__failed_test_to_edit_s} & 0.0001 \\
\metric{raw__TAB_STATE} & 0.0071 &  &  \\
\bottomrule
\end{tabular}
\end{table}
 
\begin{table}[H]
\centering
\caption{Distribution of behavioral contexts among held-out query-imminence observation windows at the reported window (30\,s), with each context's positive rate. Profiles are evaluation strata from the deployment-1--2 taxonomy models.}
\label{tab:dist-imminence}
\small
\setlength{\tabcolsep}{6pt}
\renewcommand{\arraystretch}{1.12}
\begin{tabular}{lrrr}
\toprule
\textbf{Behavioral context} & \textbf{Windows} & \textbf{Share} & \textbf{Positive rate} \\
\midrule
W2: Iterating & 1,837 & 17.0\% & 80.8\% \\
After last query & 1,787 & 16.5\% & 0.0\% \\
No query this session & 1,600 & 14.8\% & 0.0\% \\
W2 context (invalid interval) & 1,289 & 11.9\% & 21.5\% \\
W1: Oriented & 1,072 & 9.9\% & 52.6\% \\
W2: Passive & 1,061 & 9.8\% & 91.3\% \\
W2: Spinning & 607 & 5.6\% & 25.5\% \\
W2: Debugging & 566 & 5.2\% & 55.1\% \\
W1: Struggling & 436 & 4.0\% & 27.5\% \\
W1 context (excluded from population) & 327 & 3.0\% & 14.7\% \\
W1: Cold Start & 243 & 2.2\% & 82.7\% \\
\bottomrule
\end{tabular}
\end{table}
 
\begin{table}[H]
\centering
\caption{Distribution of behavioral profiles among held-out help-seeking-type queries at the reported window (15\,s), with each profile's guided rate.}
\label{tab:dist-type}
\small
\setlength{\tabcolsep}{6pt}
\renewcommand{\arraystretch}{1.12}
\begin{tabular}{lrrr}
\toprule
\textbf{Preceding profile} & \textbf{Queries} & \textbf{Share} & \textbf{Guided} \\
\midrule
W2: Passive & 186 & 37.3\% & 22.6\% \\
W2: Iterating & 166 & 33.3\% & 33.7\% \\
W1: Oriented & 47 & 9.4\% & 72.3\% \\
W2: Debugging & 28 & 5.6\% & 35.7\% \\
W2 (invalid interval) & 25 & 5.0\% & 36.0\% \\
W1: Cold Start & 20 & 4.0\% & 0.0\% \\
W2: Spinning & 13 & 2.6\% & 46.2\% \\
W1: Struggling & 10 & 2.0\% & 60.0\% \\
W1 (excluded from population) & 4 & 0.8\% & 50.0\% \\
\bottomrule
\end{tabular}
\end{table}
 
\end{document}